\documentclass[preprint,12pt]{elsarticle}

\usepackage{amssymb}
\usepackage{amsmath}
\usepackage{graphicx}
\usepackage{algorithmic}
\usepackage[ruled]{algorithm2e}
\usepackage{array}
\usepackage{textcomp}
\usepackage{stfloats}
\usepackage{url}
\usepackage{float}

\usepackage{hyperref}
\hypersetup{
    breaklinks=true,
    colorlinks=true,
    urlcolor=blue,
    citecolor=blue,
    linkcolor=blue
}
\newcommand{\cmark}{\textcolor[rgb]{0,0.55,0}{\ding{51}}}
\newcommand{\xmark}{\textcolor[rgb]{0.8,0,0}{\ding{55}}}

\usepackage{verbatim}
\usepackage[utf8]{inputenc}
\usepackage{booktabs}
\usepackage{color}
\usepackage[dvipsnames]{xcolor}
\usepackage[table]{xcolor}
\usepackage{pifont}  % For dingbats symbols
\usepackage{xr-hyper}

\usepackage{multirow}
\usepackage[normalem]{ulem}%
\usepackage{xcolor}

\newcommand{\rev}[1]{\textcolor{black}{#1}}
\newcommand{\revMi}[1]{\textcolor{black}{#1}}

\usepackage{graphicx}
\usepackage{float}
\usepackage{multirow}
\usepackage{subcaption}

\def\BibTeX{{\rm B\kern-.05em{\sc i\kern-.025em b}\kern-.08em
    T\kern-.1667em\lower.7ex\hbox{E}\kern-.125emX}}

\journal{Journal of Systems Architecture}

\begin{document}

\begin{frontmatter}



%\title{I-LW-DETR: Integer-Only Vision Transformer for Efficient Object Detection} \martyna{ classical, beside quite identical as for I-segmenter}

\title{Enabling Fully Integer-Only Inference for Lightweight Detection Transformers}

%% \author[label1]{}

\author[1,2]{Thanh Cong Le}
\ead{Thanh-Cong.LE@cea.fr}

\author[1]{Michal Szczepanski}
\ead{michal.szczepanski@cea.fr}

\author[1]{Martyna Poreba}
\ead{martyna.poreba@cea.fr}

\affiliation[1]{
    organization={Université Paris-Saclay, CEA, List},
    city={Palaiseau},
    postcode={F-91120},
    country={France}
}

\affiliation[2]{
    organization={Université d'Evry Paris-Saclay},
    city={Évry-Courcouronnes },
    postcode={F-91000},
    country={France}
}

%%% ANONONYMOUS
%\author{} %% Author name

%\affiliation{organization={},%Department and Organization

%% Abstract
\begin{abstract}
%\textcolor{red}{typically 150–300 words}\\
%Vision Transformer detectors have narrowed the accuracy gap with CNNs but remain difficult to deploy on 
Vision Transformer detectors now approach the accuracy of CNNs but remain difficult to deploy on NPUs and microcontrollers because key components, including deformable attention, feature fusion, and nonlinear activation functions, are not natively compatible with integer arithmetic. Existing quantized detectors often retain operators such as Softmax, GELU, and LayerNorm in floating point%or focus on heavyweight backbones, leaving lightweight detection transformers without an end-to-end integer implementation. %, or, as in the case of the only fully integer-only DETR reported to date, target a vanilla DETR with standard attention and a heavyweight backbone.}
Deformable, multi-scale detection transformers therefore still lack an end-to-end integer implementation. We address this gap with I-LW-DETR, the first fully integer-only lightweight DETR, in which every operation in the forward pass, including transformer nonlinearities, is executed in integer arithmetic. I-LW-DETR is built upon three key components: a scale-preserving split convolution that assigns independent activation scale to each branch of the multi-scale projector; SD-ShiftGELU, a sign-dependent GELU approximation that preserves element-wise behavior the accuracy collapse of the original ShiftGELU; and a constrained Shiftmax that maintains stable Softmax normalization. Experimental results demonstrate that the proposed quantization pipeline consistently produces efficient fully integer-only models across different model scales. Across all model scales, the proposed pipeline incurs only a moderate accuracy degradation while reducing the model size by approximately $3.6\times$ and the computational cost by more than one order of magnitude. Code will be available at \href{https://github.com/ms245755/I-LW-DETR }{github.com/ms245755/I-LW-DETR}
\end{abstract}


\end{frontmatter}


\section{Introduction}
\label{sec:Intro}
Convolutional object detectors, particularly the YOLO family \cite{yolo,yolo10,tian2025yolov, ZHAO2026103777}, remain the dominant choice for real-time vision because of their excellent speed-accuracy trade-off. In parallel, transformer-based detectors have emerged as an alternative paradigm by formulating object detection as an end-to-end set prediction problem. Built upon Vision Transformers (ViTs)~\cite{dosovitskiy2020image}, the DETR family~\cite{detr} formulates detection as direct set prediction, eliminating hand-crafted components such as anchor generation and non-maximum suppression. Subsequent research further improved deployment efficiency, with lightweight variants such as LW-DETR~\cite{lwdetr} substantially narrowing the throughput and parameter gap with convolutional detectors. Despite these advances, key transformer operators including attention, multi-scale feature projection and nonlinear activation functions, still rely on floating-point arithmetic, limiting efficient deployment on integer-only edge accelerators. Consequently, deploying accurate transformer detectors on resource-constrained edge hardware remains a major challenge.

Although quantization is the standard approach for efficient inference, an important distinction is often overlooked. Most existing transformer quantization methods primarily target linear layers while leaving operations such as Softmax, GELU, and LayerNorm in floating-point \cite{PTQ4ViT, Q8BERT, iptqvit}. In contrast, truly integer-only inference requires every operation, including the transformer non-linearities, to be implemented in integer arithmetic. Prior work has demonstrated this for transformer encoders in image classification, where I-BERT~\cite{ibert} and I-ViT~\cite{ivit} replace floating-point exponentials and divisions with shift-based integer approximations. More recently, I-Segmenter \cite{isegmenter} extended this paradigm to semantic segmentation, becoming the first fully integer-only ViT for dense prediction. Achieving fully integer-only inference for modern detection transformers is considerably more challenging because many architectures rely on deformable attention, multi-scale feature projectors with heterogeneous activation ranges, and non-linearities whose integer approximations degrade sharply under outlier activations. %Existing DETR quantization methods \sout{either} \michal{mostly} rely on partial integer execution~\cite{qdetr,aqdetr,qrtdetr} %or achieve fully integer-only inference only on heavyweight detector architectures~\cite{eiqdetr}. Fully integer-only inference for lightweight deformable DETRs therefore remains largely unexplored.
%\michal{EIQ-DETR~\cite{eiqdetr} established that a DETR detector can be executed entirely in integer arithmetic, but it targets a vanilla DETR with a Swin-T backbone, standard multi-head attention and a single-scale feature map. It does not address the components that make lightweight detectors such as LW-DETR efficient.} \martyna{ZMIEN!}

We address this gap with I-LW-DETR. Rather than applying a generic quantization pipeline to an existing detector, we jointly redesign the architecture and its integer approximations so that the entire network executes in integer arithmetic. Certain operations are inherently incompatible with integer-only hardware and therefore require architectural redesign rather than numerical approximation alone. 
The remaining challenges arise from numerical precision. Straightforward integer quantization reduces projector activation resolution and degrades the accuracy of transformer non-linearities under outlier activations. Together, these architectural and numerical improvements enable I-LW-DETR to preserve detection accuracy under fully integer-only inference while substantially reducing model size and computational cost compared with existing integer DETR detectors.

Our contributions are threefold:
\begin{itemize}
    %\item We present I-LW-DETR, a fully integer-only lightweight detection transformer that extends integer-only inference to deformable-attention-based architectures.
    \item \rev{We present I-LW-DETR, to the best of our knowledge the first fully integer-only lightweight DETR with deformable attention and multi-scale feature fusion;
    }
    
    %\item We introduce three integer approximation blocks that resolve the main architectural and numerical obstacles to integer-only transformer detection while preserving detection accuracy.
    
    \item \rev{We identify three key limitations of existing integer operators for deformable DETRs and address them with tailored integer-exact solutions: a scale-preserving split convolution, SD-ShiftGELU, and Constrained Shiftmax;}

    \item We validate the proposed framework under both PTQ and QAT, demonstrating state-of-the-art performance while substantially reducing model size and computational cost.

\end{itemize}

This paper is organized as follows. Section~\ref{sec:related} reviews related work on object detection and the quantization of DETR-based object detectors. We then introduce, in Section~\ref{sec:methodology}, I-LW-DETR and its integer-only quantization pipeline. Experimental settings and results, and ablation studies are presented in Sections~\ref{sec:setup}, \ref{sec:results}, and~\ref{sec:ablation}, respectively. \rev{Section~\ref{sec:discussion} discusses the main findings, hardware compatibility, and limitations of the proposed approach.} %Section~\ref{sec:mappability} discusses the hardware mappability of the proposed operators to integer edge accelerators. 
Finally, Section~\ref{sec:conclusions} concludes the paper.  % version finale
\section{Related Work}
\label{sec:related}

\subsection{DETR-Based Object Detection}

DETR \cite{detr} reformulates object detection as direct set prediction, pairing a transformer encoder-decoder together with bipartite Hungarian matching. This end-to-end formulation eliminates hand-crafted components such as anchor boxes and non-maximum suppression while simplifying the detection pipeline. However, these advantages come at the cost of slow convergence and a substantial computational footprint. Subsequent work targeted these limitations. Deformable DETR \cite{deformabledetr} replaced dense global attention with deformable attention, in which each query attends only to a few sampled locations around a reference point, accelerating convergence and reducing cost. DINO \cite{zhang2022dino} improved query initialization and denoising training, and RT-DETR \cite{rtdetr, rtdetrv2} introduced a hybrid encoder and uncertainty-aware query selection for real-time inference. More recent work such as D-FINE~\cite{peng2024dfine} further improves the accuracy--efficiency trade-off by reformulating bounding-box regression as fine-grained distribution refinement. 
\rev{Recent DETR variants have also targeted the specific challenges of small-object detection in aerial imagery. UAV-DETR~\cite{uavdetr} enhances multi-scale representations through spatial--frequency feature fusion, frequency-aware down-sampling, and semantic alignment. SO-DETR~\cite{sodetr} further combines a dual-domain hybrid encoder with enhanced query selection and knowledge distillation to improve efficient small-object detection, while FA-DETR~\cite{fadetr} strengthens fine-grained feature preservation and localization through edge enhancement, learnable positional encoding, and multi-domain feature fusion.} \rev{In parallel,} LW-DETR~\cite{lwdetr} pursues a lightweight design built on a plain ViT encoder that jointly performs feature extraction and interaction, combining global, window, and deformable attention to balance accuracy and efficiency. Building on this design, RF-DETR~\cite{robinson2026rfdetr} integrates neural architecture search to identify more efficient model configurations.

\subsection{Integer Quantization of Vision Transformers}
Compared with convolutional networks, transformers are considerably more difficult to quantize because LayerNorm, Softmax, and GELU exhibit highly non-uniform activation distributions that are poorly suited to low-precision integer arithmetic. Most post-training quantization (PTQ) methods address these challenges through operator-specific approximations. PSAQ-ViT \cite{psaqvit} enables data-free quantization by synthesizing calibration samples through a patch-similarity metric. FQ-ViT \cite{fqvit} introduces a power-of-two factor for the inter-channel variation of LayerNorm inputs and a log-int-softmax that quantizes attention with bit-shift arithmetic. RepQ-ViT \cite{repqvit} reparameterizes quantization scales to make the LayerNorm and Softmax distributions more amenable to low-bit inference. Another line of research addresses activation outliers directly. For instance, SPIQ \cite{spiq} assigns a separate quantization scale to each input channel when different features exhibit different activation ranges. More recently, ORQ-ViT \cite{HE2025103530} introduced an outlier-resilient strategy based on outlier decomposition, reducing the adverse effect of extreme activation values while retaining the efficiency of PTQ.

A second research direction focuses on end-to-end integer-only inference. I-BERT \cite{ibert} approximates transformer non-linearities entirely in integer arithmetic for NLP, and I-ViT \cite{ivit} extends this to ViTs with the bit-shift-based ShiftExp, Shiftmax, and ShiftGELU operators over a dyadic integer pipeline. More recently, I-Segmenter \cite{isegmenter} realizes the first fully integer-only ViT for semantic segmentation, introducing a $\lambda$-ShiftGELU activation for the long-tailed distributions that destabilize quantized dense prediction. However, none of these methods tackles the additional architectural and numerical challenges posed by lightweight DETR-based object detection.

\subsection{Quantization of DETR Models}
A smaller body of work focuses specifically on DETR detectors. Q-DETR \cite{qdetr} proposes the first quantization-aware training framework for DETR, identifying query information distortion as the main source of accuracy loss and correcting it through a distribution-rectification knowledge-distillation scheme. AQ-DETR \cite{aqdetr} introduces auxiliary queries with a layer-by-layer distillation strategy to preserve the representational capacity of quantized queries, enabling stable low-bit (down to 4-bit) detection. QRT-DETR \cite{qrtdetr} develops a post-training quantization (PTQ) pipeline for the real-time RT-DETR, combining an EMA-MSE quantizer with reconstruction-aware optimization to mitigate the severe accuracy degradation observed at low bit widths. \rev{Reg-PTQ~\cite{regptq} introduces regression-specialized calibration and quantization to better preserve bounding-box regression under low precision, and evaluates the approach on DETR in addition to CNN-based detectors. IGQ-ViT~\cite{igqvit} applies instance-aware group quantization to transformer activations and also reports results on DETR. More recently, QwT~\cite{qwt} introduces lightweight compensation modules with closed-form calibration and demonstrates its applicability to DETR-based object detection.} %However, these methods reduce the precision of weights and/or activations while leaving transformer non-linearities in floating-point, preventing fully integer-only inference.
\rev{However, these methods primarily focus on low-bit quantization of weights and activations and do not provide fully integer-only inference across the complete DETR computational graph. While EIQ-DETR~\cite{eiqdetr} demonstrated fully integer-only inference for a conventional DETR architecture, this problem remains largely unexplored for lightweight deformable DETRs.}
%EIQ-DETR \cite{eiqdetr} achieves fully integer-only DETR inference by extending integer approximation to all non-linear layers, including Sigmoid, Softmax, LayerNorm, and GELU, but relies on a comparatively large Swin-T backbone, several times the size and computational cost of the lightweight regime.
%\michal{EIQ-DETR \cite{eiqdetr} was the first to achieve fully integer-only DETR inference, extending integer approximation to all non-linear layers, including Sigmoid, Softmax, LayerNorm, and GELU. It quantizes a vanilla DETR with a Swin-T backbone, standard multi-head attention and a single-scale feature map. %The resulting model is several times larger and more computationally expensive than lightweight detectors, and it contains none of their characteristic components, namely deformable cross-attention, multi-scale feature projection and two-stage query selection.}
  % version finale
\section{Methodology}
\label{sec:methodology}

I-LW-DETR is a fully integer-only realization of LW-DETR~\cite{lwdetr}. Rather than directly quantizing the original architecture, we redesign selected components to eliminate floating-point operations while preserving the lightweight architecture. The proposed framework combines quantization-ready architectural adaptations with dedicated integer formulations of transformer non-linearities, resulting in a fully integer-only inference pipeline.

\begin{figure}[!t]
    \centering
    \includegraphics[width=\linewidth]{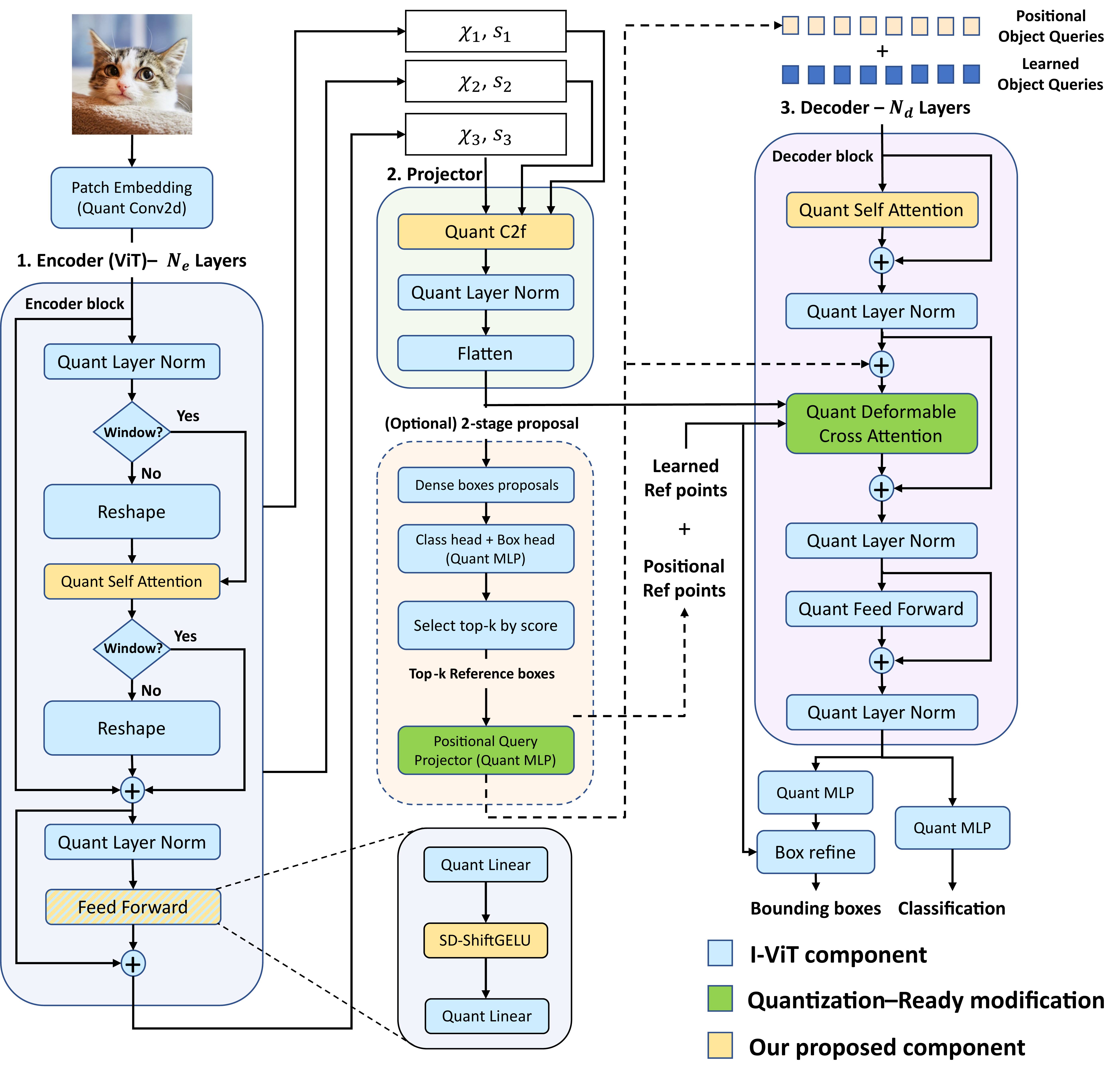}
    \caption{Overview of the proposed quantized I-LW-DETR pipeline.}
    \label{fig:quantized_lwdetr_overview}
\end{figure}

Figure~\ref{fig:quantized_lwdetr_overview} summarizes the proposed framework. Blue blocks denote integer operators adopted from I-ViT \cite{ivit}. The green blocks correspond to the architectural adaptations required for integer-only execution, while yellow blocks highlight the new quantization optimizations introduced in this work.

\subsection{Quantization-Ready Adaptation of LW-DETR}
\label{sec:qrdetr}

We first construct QR-LW-DETR, a quantization-ready floating-point variant of LW-DETR. It keeps the main architecture unchanged, but replaces two components that are not naturally compatible with integer-only inference: bilinear interpolation in deformable attention, which relies on fractional interpolation weights, and the sinusoidal encoding of reference boxes used for positional query generation, which requires trigonometric functions. QR-LW-DETR then serves as the floating-point baseline from which the final integer-only I-LW-DETR models are quantized.

\subsubsection{Integer-Compatible Deformable Attention Sampling}

The decoder of LW-DETR employs deformable attention for cross-attention between object queries and image features. Instead of attending to all spatial tokens, each query attends only to a small set of sampled feature locations around a reference point. These sampling locations are predicted by the model and are generally continuous rather than aligned with the discrete feature-map grid. Consequently, the corresponding feature values are estimated by bilinear interpolation from the four neighboring grid points. The interpolation weights depend on the relative position of the sampling point. Although bilinear interpolation provides sub-pixel sampling accuracy, it relies on fractional interpolation weights and multiple floating-point multiply-accumulate operations, making efficient integer-only implementation difficult. 

To simplify this component, bilinear interpolation is replaced by nearest-neighbor sampling. Instead of computing a weighted combination of neighboring pixels, the sampled feature is taken directly from the nearest grid point. This removes the interpolation coefficients and the associated floating-point computations while leaving the overall deformable attention mechanism unchanged. 

\subsubsection{Learnable Positional Query Generation}

LW-DETR adopts a two-stage detection pipeline in which the encoder first selects the top-\(K\) reference boxes. These reference boxes are then converted into positional queries through sinusoidal positional encoding before being processed by the decoder. While effective in floating-point inference, sinusoidal positional encoding relies on trigonometric functions that are inefficient for integer-only hardware and typically require dedicated approximations or lookup tables.

Inspired by learnable positional representations, we replace the sine/cosine reference-box encoding with a learnable projection module. Instead of computing trigonometric functions from the reference box coordinates, the module learns to transform these coordinates into positional object-query features using linear layers. These layers can be quantized using the same integer matrix-multiplication pipeline as the other linear layers.

\subsection{Integer Quantization of Linear Operations}
\label{splitconv}

\subsubsection{Standard Linear Operators}

The linear operators of I-LW-DETR (projections, convolutions, and the attention matrix multiplications) follow the standard integer-only formulation of I-ViT~\cite{ivit}. Under symmetric quantization an activation and weight tensor are represented as :
\begin{equation}
X \approx S_x I_x, \qquad W \approx S_w I_w,
\label{eq:sym_quant}
\end{equation}
where $I_x$ and $I_w$ denote the integer tensors and $S_x$ and $S_w$ their corresponding scaling factors. A linear layer
\begin{equation}
Y = XW + b
\label{eq:linear_fp}
\end{equation}
is therefore approximated as 

\begin{equation}
Y \approx S_x S_w \,(I_x I_w) + b,
\label{eq:linear_quant}
\end{equation}

where the multiply-accumulate operation is performed entirely in the integer domain and the scaling factors are applied after accumulation. Weights use symmetric per-channel quantization, whereas activations use symmetric per-tensor quantization, providing a good trade-off between numerical precision and implementation simplicity.

\subsubsection{Scale-Preserving Split Convolution}

\begin{figure}[!t]
    \centering
    \subfloat[\footnotesize Original projector convolution with a shared activation scale.]{
        \includegraphics[width=0.95\linewidth]{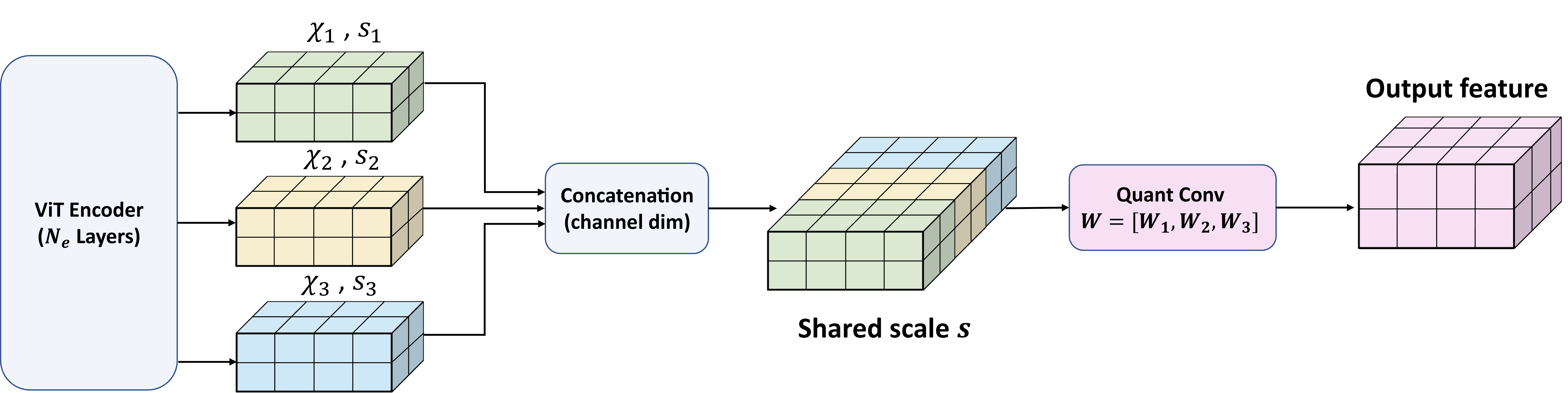}
        \label{fig:shared_scale_conv}
    }
    \vspace{0.5em}
    \subfloat[\footnotesize Proposed scale-preserving split convolution.]{
        \includegraphics[width=0.95\linewidth]{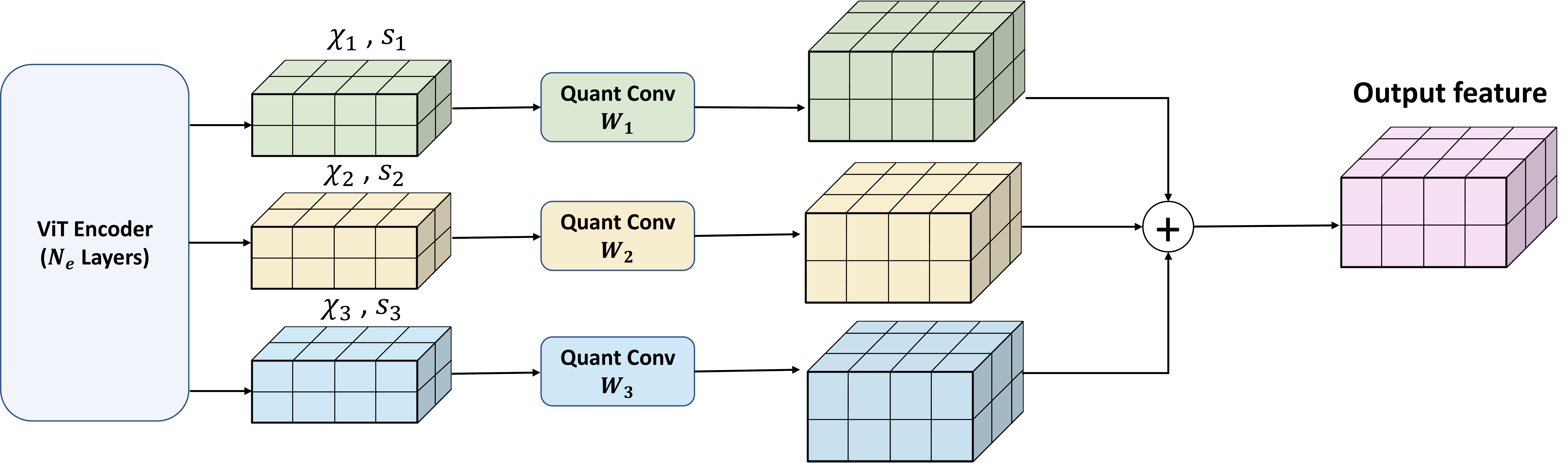}
        \label{fig:split_scale_conv}
    }
    \caption{Projector implementation before and after the proposed reformulation. }
    \label{fig:projector_split_conv}
\end{figure}

A limitation of this quantization scheme appears in the projector of LW-DETR. The projector receives multiple feature maps from different stages of the ViT encoder, which typically exhibit markedly different activation ranges. Under the I-ViT quantization framework, however, all concatenated feature maps share a single activation scale. As a result, the scale is determined by the largest activation values, reducing the effective quantization resolution of feature maps with smaller dynamic ranges, as illustrated in Figure~\ref{fig:projector_split_conv}(a).

To address this issue, we reformulate the projector convolution into multiple independent convolution branches, allowing each encoder output to retain its own activation scale while remaining compatible with the I-ViT quantization framework, as illustrated in Figure~\ref{fig:projector_split_conv}(b). Because the branches use different activation scales, their partial integer outputs cannot be accumulated directly. Before accumulation, each branch is therefore rescaled to a common output scale. We choose this scale as the maximum of the calibrated per-branch scales. Since the scale ratios are fixed after calibration, the rescaling reduces to a constant affine transformation that is implemented as an integer dyadic multiply-and-shift during inference, fully consistent with the I-ViT pipeline. Consequently, the accumulation remains entirely in the integer domain.

\subsection{Integer Approximation of Nonlinear Operations}
\label{sec:nonlinear}

Integer implementations of transformer non-linearities have previously been introduced in I-ViT. We adopt its LayerNorm formulation, in which the variance is normalized through an integer square root computed by Newton's method. To improve the numerical stability of the decoder, whose activations span a wider dynamic range than the encoder, we increase the number of Newton iterations from 10 to 20. The remaining nonlinear operators require additional adaptations for I-LW-DETR. In the following subsections, we introduce modified integer formulations of Softmax, GELU, and SiLU that improve robustness to the wider activation distributions encountered during object detection.

\subsubsection{Sign-Dependent ShiftGELU}
\label{sec:sdshiftgelu}

GELU is used in the feed-forward blocks of the ViT encoder. Following I-ViT~\cite{ivit}, it is approximated by
\begin{equation}
\label{eq:gelu}
\mathrm{GELU}(x)\approx x\,\sigma(1.702x),
\end{equation}
where the factor \(1.702\) is implemented using additions and bit shifts. Let \(x\approx S_xI_{\mathrm{in}}\) denote the quantized input. The corresponding integer sigmoid argument is
\begin{equation}
I_p=
I_{\mathrm{in}}
+
(I_{\mathrm{in}}\gg1)
+
(I_{\mathrm{in}}\gg3)
+
(I_{\mathrm{in}}\gg4),
\end{equation}
% and the exponential term is evaluated using the ShiftExp operator introduced in the previous subsection.
The exponential term in the sigmoid approximation is evaluated using the ShiftExp operator introduced in I-ViT. For a quantized input \(x \approx S I\), we write this operation as
\begin{equation}
S_{\exp} I_{\exp}
=
\operatorname{ShiftExp}(S, I)
\approx
\exp(SI),
\end{equation}
where \(I_{\exp}\) is the integer output and \(S_{\exp}\) is its associated scaling factor. ShiftExp makes the exponential compatible with integer-only inference by approximating it with integer additions, floor operations, and bit shifts.

To make the input to ShiftExp \sout{is} non-positive, the ShiftGELU implementation from I-ViT applies a vector-wise maximum subtraction before the exponential approximation:

\begin{equation}
I_{\Delta} = I_p - \max(I_p).
\end{equation}

However, this operation introduces a dependency between elements of the same vector. If one element has a large positive value, it becomes the maximum and shifts all other elements toward more negative values.

\begin{figure}[!h]
    \centering
    \subfloat[{\footnotesize ShiftExp vs \(\exp(x)\).}]{
        \includegraphics[width=0.445\linewidth]{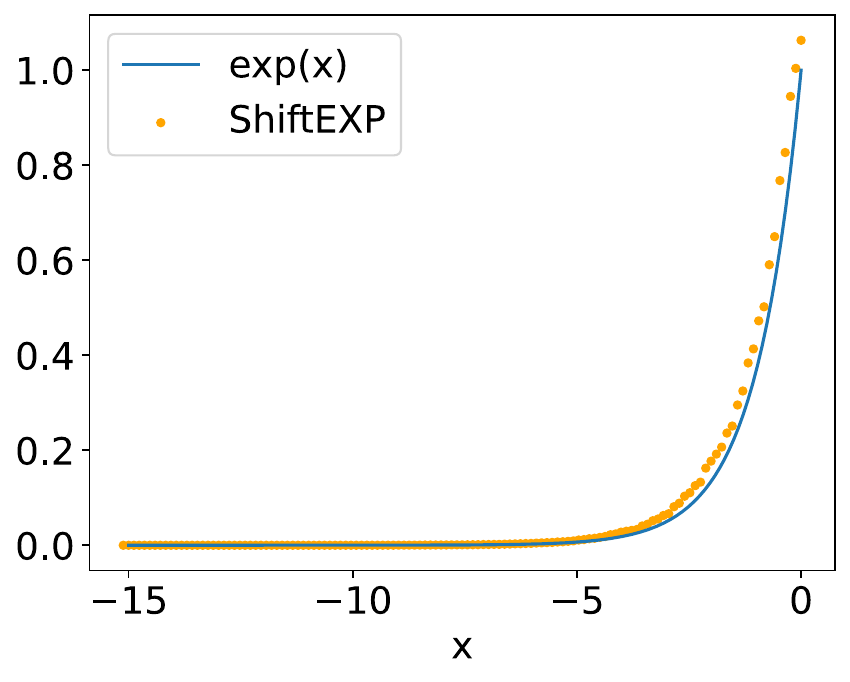}
        \label{fig:exp_vs_intexp}
    }
    \hfill
    \subfloat[{\footnotesize Relative error of ShiftExp.}]{
        \includegraphics[width=0.46\linewidth]{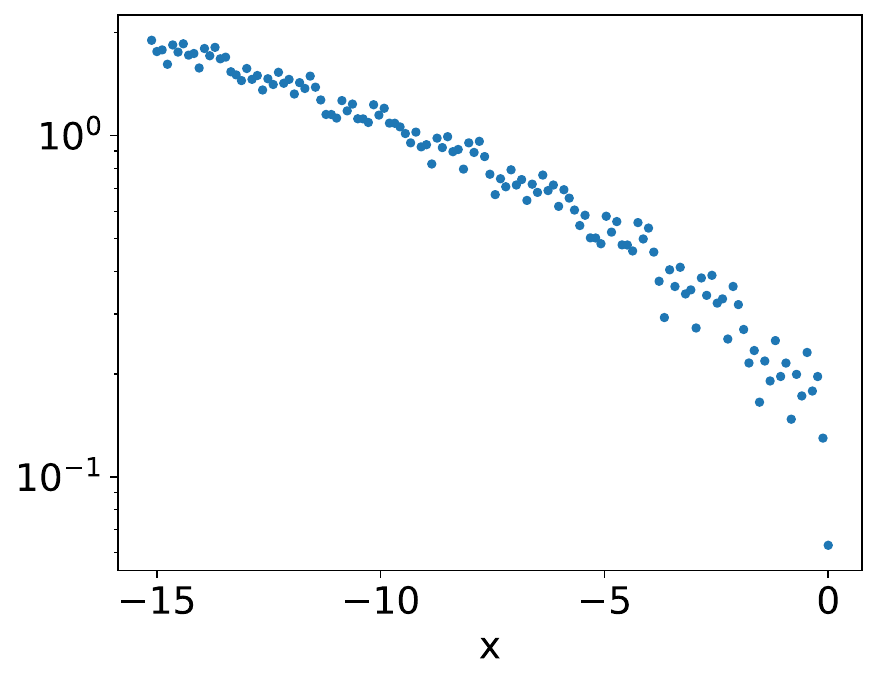}
        \label{fig:intexp_relative_error}
    }
    \caption{Numerical behavior of the ShiftExp approximation on negative inputs.}
    \label{fig:shiftexp_analysis}
\end{figure}

This behavior is problematic because ShiftExp is sensitive to large negative inputs. As shown in Figure~\ref{fig:shiftexp_analysis}, ShiftExp follows the floating-point exponential on negative inputs, but its relative approximation error increases as the input becomes more negative. Therefore, the vector-wise maximum subtraction can push many activations into a high-error region of ShiftExp, degrading the GELU approximation.

To address this limitation,
%To remove this dependency, 
we propose Sign-Dependent ShiftGELU (SD-ShiftGELU). Instead of relying on a vector-wise maximum subtraction, the sigmoid function is reformulated as

\begin{equation}
\sigma(z)
=
\begin{cases}
\dfrac{1}{1+\exp(-z)}, & z \geq 0, \\[6pt]
\dfrac{\exp(z)}{1+\exp(z)}, & z < 0.
\end{cases}
\end{equation}

\begin{figure}[!htbp]
    \centering
    \includegraphics[height=0.15\textheight]{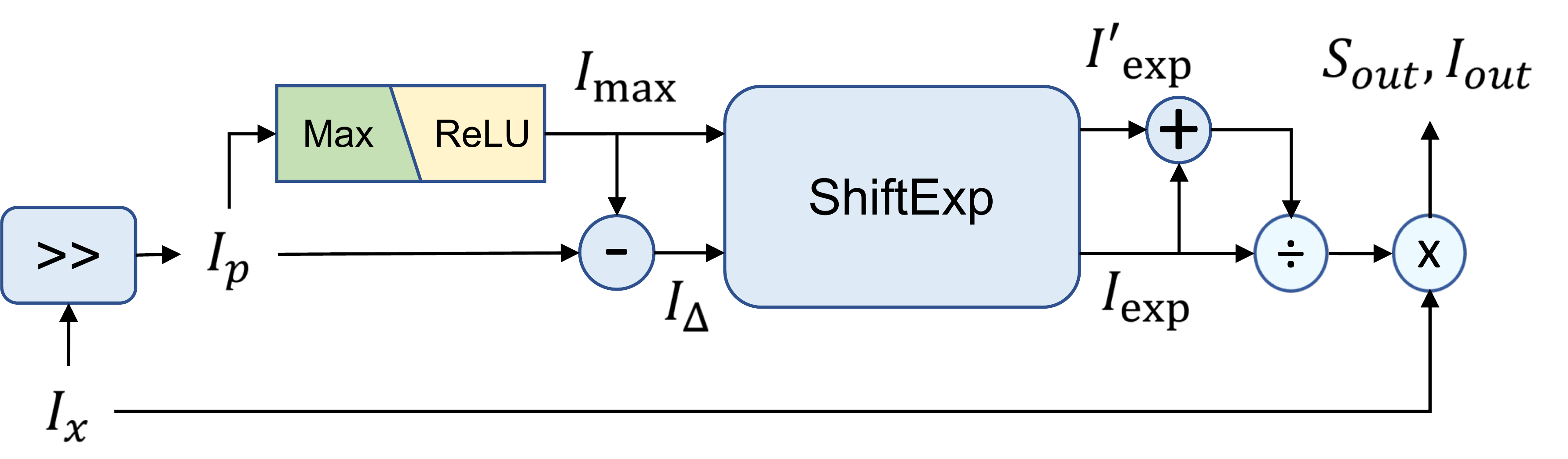}
    \caption{Overview of the ShiftGELU variants. The green block corresponds to the original I-ViT formulation, while the yellow block corresponds to the proposed sign-dependent formulation used in our method.}
    \label{fig:shiftgelu_comparison}
\end{figure}

The two expressions are mathematically equivalent while ensuring %but ensure 
that the exponential input remains non-positive in both cases. 

Each activation is processed independently according to its sign, eliminating the need for a global maximum operation (Figure~\ref{fig:shiftgelu_comparison}). The corresponding integer input to ShiftExp becomes

\begin{equation}
I_{\Delta}
=
I_p - \mathrm{ReLU}(I_p),
\end{equation}

which guarantees \(I_{\Delta} \leq 0\) for every element independently.

Unlike the original ShiftGELU, the proposed formulation preserves the element-wise behavior of GELU while avoiding unnecessary shifts toward the high-error region of ShiftExp.

\subsubsection{Constrained Shiftmax}
\label{sub:shiftmax}

I-ViT Shiftmax performs integer Softmax normalization using a fixed precision parameter \(M\). 
In I-LW-DETR, however, the accumulated exponential sum can vary significantly across attention layers and may exceed the effective precision supported by 32-bit integer arithmetic. As a result, the normalization factor becomes too coarse and may even collapse to zero. 

To address this limitation, we introduce an adaptive Constrained Shiftmax that calibrates a denominator shift \(s_d\) for each Shiftmax block. 

Let \(I_{\exp_i}\) denote the integer exponential value produced by ShiftExp for the \(i\)-th attention score. The original I-ViT Shiftmax computes the integer Softmax output as
\begin{equation}
I_{\mathrm{out}_i}
=
\left(
\left\lfloor
\frac{2^M - 1}
{\sum_{j=1}^{d} I_{\exp_j}}
\right\rfloor
I_{\exp_i}
\right)
\gg
\left(
M-(k_{\mathrm{out}}-1)
\right),
\label{eq:shiftmax_original}
\end{equation}
where \(k_{\mathrm{out}}\) is the output bit-width and \(M\) is the integer-division precision. Since values above 31 exceed the native precision of 32-bit integer arithmetic, we fix \(M = M_{\mathrm{hw}} = 31\).

Formally, the accumulated exponential sum is defined as

\begin{equation}
S_{\mathrm{sum}}
=
\sum_{j=1}^{d} I_{\exp_j}.
\end{equation}
For large values of \(S_{\mathrm{sum}}\), the normalization factor cannot be represented accurately within the fixed hardware precision and may collapse to zero.

The shifted normalization is expressed as

\begin{equation}
I_{\mathrm{out}_i}
=
\left(
\left\lfloor
\frac{2^M - 1}
{
\sum_j (I_{\exp_j} \gg s_d)
}
\right\rfloor
\cdot
\left(I_{\exp_i} \gg s_d\right)
\right)
\gg
\left(
M-(k_{\mathrm{out}}-1)
\right).
\label{eq:shiftmax_shifted}
\end{equation}

The denominator shift adaptively reduces the dynamic range of the accumulated exponential sum, allowing the normalization to remain representable within the fixed hardware precision.

The choice of \(s_d\) controls the trade-off between division precision and information preservation. If \(s_d\) is too small, the normalization factor remains poorly represented and may even collapse to zero. Conversely, an excessively large shift discards small exponential values $I_{\exp_i}$ and weak attention contributions. Since the exponential-sum distribution varies across Shiftmax blocks, a fixed denominator shift is suboptimal.

To select \(s_d\) automatically, we observe the exponential-sum statistics of each Shiftmax block during the calibration process. The required precision is estimated as
\begin{equation}
M_{\mathrm{req}}
=
\left\lceil
\log_2
\left(
S_{\mathrm{sum}}
\right)
\right\rceil,
\qquad
M_{\mathrm{target}}
=
M_{\mathrm{req}} + m_p ,
\end{equation}
where \(m_p\) is an additional precision margin for the integer division. The denominator shift is then computed as
\begin{equation}
s_d
=
\max
\left(
M_{\mathrm{target}} - M_{\mathrm{hw}},
0
\right).
\end{equation}
In this work, \(m_p=4\) is used as a fixed safety margin to improve normalization precision, and the same value is applied to all Shiftmax blocks and model sizes.

A naive implementation of the shifted denominator would compute
\begin{equation}
S_R^{\mathrm{naive}}
=
\sum_i
\left(
I_{\exp_i} \gg s_d
\right),
\end{equation}
which discards the remainder of each exponential term. To reduce this loss, each exponential value is decomposed into quotient and remainder:
\begin{equation}
I_{\exp_i}
=
Q_i2^{s_d}+r_i,
\qquad
Q_i=
\left\lfloor
\frac{I_{\exp_i}}
{2^{s_d}}
\right\rfloor .
\end{equation}
The reduced denominator is then computed as
\begin{equation}
S_R
=
\sum_i Q_i
+
\left\lfloor
\frac{\sum_i r_i}
{2^{s_d}}
\right\rfloor
=
\left\lfloor
\frac{S_{\mathrm{sum}}}
{2^{s_d}}
\right\rfloor .
\end{equation}
This quotient-remainder accumulation recovers the exact shifted sum, instead of summing individually rounded terms.

Finally, the shifted denominator is constrained to the 32-bit range and used to compute the normalization factor:
\begin{equation}
S_R
=
\mathrm{clamp}
\left(
S_R,
1,
2^{31}-1
\right),
\qquad
F
=
\left\lfloor
\frac{2^{31}-1}
{S_R}
\right\rfloor .
\end{equation}
The integer Softmax output is obtained as
\begin{equation}
I_{\mathrm{out}_i}
=
\left\lfloor I_{\exp_i}F \right\rfloor \gg (31-k_{\mathrm{out}}+1+s_d)
\end{equation}
  % version finale
\section{Experimental Setup}
\label{sec:setup}

\subsection{Datasets and Evaluation Metrics}
COCO ~\cite{lin2015microsoftcococommonobjects} is the primary benchmark used in this work. It contains 80 object categories, and all models are evaluated following the standard COCO evaluation protocol. The primary metric is mean Average Precision (mAP), averaged over ten IoU thresholds from 0.50 to 0.95 with a step size of 0.05. We additionally report \(\mathrm{AP}_S\), \(\mathrm{AP}_M\), and \(\mathrm{AP}_L\) to assess detection performance on small, medium, and large objects, respectively. As a more challenging evaluation, we consider VisDrone~\cite{zhu2021detection}, an aerial object detection benchmark characterized by numerous small, densely packed, and partially occluded objects.  We use the official training and test-dev splits, and follow the standard COCO evaluation protocol using the same metrics. Unless otherwise stated, all input images are resized to \(640 \times 640\).

\subsection{Computational Cost Metrics}
\label{sec:computational_cost_metrics}

In addition to detection accuracy, we report model size and bit-operations (BOPs). Model size in MB quantifies the memory required to store the model parameters, while BOPs provide a hardware-independent estimate of arithmetic complexity by accounting for both the number of operations and the numerical precision. Following the formulation of Bayesian Bits~\cite{vanbaalen2020bayesianbitsunifyingquantization}, BOPs are computed as

\begin{equation}
\mathrm{BOPs}
=
\sum_{l}
\mathrm{Ops}_{l}
\times
b_{\mathrm{a},l}
\times
b_{\mathrm{w},l},
\end{equation}

\revMi{where \(\mathrm{Ops}_{l}\) denotes the number of operations in layer \(l\), counting each multiply–accumulate (MAC) as one operation, and \(b_{\mathrm{a},l}\) and \(b_{\mathrm{w},l}\) denote the corresponding operand bit-widths. For FP32 models, both are set to 32 bits. For integer-only models, the reported BOPs also account for the additional integer operations introduced by nonlinear approximations and intermediate computations, using their respective operand precisions. Results are expressed in TBOPs (\(10^{12}\) bit operations), while operation counts are reported in GOPs (\(10^9\) operations).}
%\michal{where \(\mathrm{Ops}_{l}\) is the operation count of layer \(l\), with one multiply--accumulate operation counted as one operation, and \(b_{\mathrm{a},l}\) and \(b_{\mathrm{w},l}\) are the corresponding activation and weight bit-widths. For FP32 models, both bit-widths are set to 32. For the integer models, \(\mathrm{Ops}_{l}\) also includes the integer operations introduced by the non-linear approximations, and operations executed with 16-bit intermediate precision are counted with their corresponding bit-widths, the resulting costs being summed over the model. All BOPs are reported in TBOPs (\(10^{12}\) bit operations), and operation counts in GOPs (\(10^{9}\).}

\subsection{Model Variants}

Table~\ref{tab:lwdetr_variants} summarizes the three QR-LW-DETR variants (Section~\ref{sec:qrdetr}); they share the same general architecture but differ in encoder width, depth, query count, and computational cost.

\begin{table}[H]
\centering
\scriptsize
\setlength{\tabcolsep}{4pt}
\caption{QR-LW-DETR model variants used in the experiments.}
\label{tab:lwdetr_variants}
% \footnotesize
\setlength{\tabcolsep}{2.8pt}
\renewcommand{\arraystretch}{1.1}
\begin{tabular}{|l|ccccc|ccc|cc|}
\hline
\multirow{2}{*}{Model} 
& \multicolumn{5}{c|}{ViT Encoder} 
& \multicolumn{3}{c|}{Projector} 
& \multicolumn{2}{c|}{DETR Decoder} \\
\cline{2-11}
& Layers 
& Dim 
& \begin{tabular}{c}Global\\Attn.\end{tabular}
& \begin{tabular}{c}Window\\Attn.\end{tabular}
& \begin{tabular}{c}Output\\Layers\end{tabular}
& Blocks 
& Dim 
& Scales 
& Layers 
& Queries \\
\hline
Tiny (ViT-T)   
& 6  & 192 & 3 & 3 & 1, 3, 5    & 1 & 256 & \(1/16\) 
& 3 & 100 \\
Small (ViT-S) 
& 10 & 192 & 4 & 6 & 2, 4, 5, 9 & 1 & 256 & \(1/16\) 
& 3 & 300 \\
Medium (ViT-M)
& 10 & 384 & 4 & 6 & 2, 4, 5, 9 & 1 & 256 & \(1/16\) 
& 3 & 300 \\
\hline
\end{tabular}
\end{table}

\subsection{Training and Quantization Protocol}

All quantized models use an \(8/8/16\) quantization configuration, corresponding to 8-bit weights, 8-bit activations, and 16-bit intermediate precision for selected non-linear operations. To evaluate the proposed integer-only formulation, we consider both post-training quantization (PTQ) and quantization-aware training (QAT). All quantized models use symmetric quantization for weights and activations, where floating-point values are represented by signed integers and scaling factors without an explicit zero-point. 

For COCO, we use the original pretrained LW-DETR checkpoints as the floating-point baselines. For VisDrone, the corresponding COCO-pretrained checkpoint is adapted by replacing the classification head and fine-tuning the model on VisDrone before quantization. The new head is first trained for 5 epochs with the remaining parameters frozen. The full model is then fine-tuned for 200 epochs using a delayed cosine learning-rate schedule, where the learning rate is kept at \(10^{-4}\) until epoch 40 and then gradually decayed to \(10^{-6}\). After obtaining the floating-point baseline for each dataset, the quantization-ready architectural modifications are applied to obtain QR-LW-DETR. 

For PTQ, the pretrained FP32 weights remain fixed, and the original modules are replaced with their quantized counterparts. The calibration stage also initializes the parameters required by the proposed integer approximations, including the denominator shift \(s_d\) of each Shiftmax module. Activation quantization ranges are calibrated using 500 batches (batch size 1) with moving-average statistics and percentile-based range estimation. Unless otherwise specified, the component-wise ablation experiments are conducted under the PTQ setting by quantizing one operator type at a time.

For QAT, training starts from the calibrated PTQ model, with fake-quantization modules enabled during the forward pass. The backward pass is performed in floating-point, while gradients through rounding and flooring operations are approximated using the straight-through estimator. The calibrated model is fine-tuned for 15 epochs using a learning rate of \(5 \times 10^{-7}\). The activation observers remain frozen during QAT because the quantization ranges have already been determined during PTQ calibration. All experiments are implemented in PyTorch and conducted on four NVIDIA A100 GPUs, each with 80~GB of memory.  % version finale
\section{Results}
\label{sec:results}

\subsection{Evaluation of the Proposed Quantization Pipeline}

% ============================ COCO: MODEL ANALYSIS ============================
\begin{table*}[ht]
\centering
\scriptsize
\setlength{\tabcolsep}{4pt}
\renewcommand{\arraystretch}{1.15}
\caption{I-LW-DETR across the pipeline on COCO: floating-point baseline, quantization-ready version, and final integer models under PTQ and QAT.}
\label{tab:coco_analysis}
\begin{tabular}{llcccccccc}
\toprule
\textbf{Model} & \textbf{Setting} & \textbf{Prec.} & \textbf{Int-only} & \textbf{Size(MB)} & \textbf{BOPs(T)} & \textbf{mAP} & \textbf{AP\textsubscript{S}} & \textbf{AP\textsubscript{M}} & \textbf{AP\textsubscript{L}} \\
\midrule

 & LW-DETR & FP32   & \xmark & 45.98 & 11.52 & 42.6 & 22.7 & 47.4 & 60.0 \\
 & QR-LW-DETR       & FP32   & \xmark & 46.49 & 11.54 & 42.0 & 22.1 & 46.5 & 58.8 \\
\rowcolor{cyan!10}
 & PTQ      & 8/8/16 & \cmark & 12.85 & 0.89  & 36.0 & 17.7 & 40.5 & 51.5 \\
\rowcolor{cyan!22}
\multirow{-4}{*}{Tiny}
 & QAT      & 8/8/16 & \cmark & 12.85 & 0.89  & 37.9 & 19 & 42.1 & 54.5 \\

\midrule

 & LW-DETR & FP32   & \xmark & 55.54 & 17.06 & 48.0 & 26.8 & 52.5 & 65.6 \\
 & QR-LW-DETR       & FP32   & \xmark & 56.05 & 17.10 & 46.4 & 25.3 & 50.8 & 64.0 \\
\rowcolor{cyan!10}
 & PTQ      & 8/8/16 & \cmark & 17.21 & 1.32  & 41.6 & 21.7 & 46.5 & 57.4 \\
\rowcolor{cyan!22}
\multirow{-4}{*}{Small}
 & QAT      & 8/8/16 & \cmark & 17.21 & 1.32  & 43.7 & 23.6 & 47.8 & 60.1 \\

\midrule

 & LW-DETR & FP32   & \xmark & 107.73 & 43.84 & 52.5 & 32.7 & 57.6 & 70.6 \\
 & QR-LW-DETR & FP32   & \xmark & 108.23 & 43.88 & 50.8 & 30.2 & 55.5 & 68.8 \\
\rowcolor{cyan!10}
 & PTQ      & 8/8/16 & \cmark & 30.44 & 3.04  & 44 & 24.9 & 48.4 & 60.6 \\
\rowcolor{cyan!22}
\multirow{-4}{*}{Medium}
 & QAT      & 8/8/16 & \cmark & 30.44 & 3.04  & 47.1 & 26.5 & 51.5 & 64.6 \\

\bottomrule
\end{tabular}
\end{table*}

Applying PTQ drastically reduces the deployment cost, decreasing the model size by approximately $3.6\times$ and the BOPs by more than one order of magnitude for all three model variants. This substantial reduction is accompanied by a corresponding loss in detection accuracy, highlighting the challenge of fully integer-only inference without retraining.
QAT effectively compensates for this degradation without increasing the deployment cost, consistently recovering between 1.9 and 3.1 mAP across the three model variants. These results demonstrate that fine-tuning under simulated quantization substantially improves the robustness of the integer-only models while preserving the efficiency gains achieved by PTQ.
%\sout{A consistent trend across all experiments is that quantization robustness improves with model capacity. While all three variants benefit from QAT, the Medium model retains the largest fraction of its floating-point performance. Its accuracy improves from 44 mAP after PTQ to 47.1 mAP after QAT, compared with a floating-point baseline of 50.8 mAP. These results indicate that larger models are more resilient to integer quantization, consistently preserving a larger fraction of their floating-point accuracy after QAT.}
\revMi{Quantization robustness is not monotonic in model size. After QAT, the Tiny, Small and Medium models retain 90.2\%, 94.2\% and 92.7\% of the mAP of their QR-LW-DETR baselines, respectively, compared with 85.7\%, 89.7\% and 86.6\% after PTQ. The Tiny is thus the most sensitive to integer quantization and the Small the most robust.% an ordering that is also observed on VisDrone (Table~\ref{tab:visdrone_results}).
}

\revMi{The practical benefit of integer-only inference is best assessed at a fixed hardware budget rather than by comparing each model with its own floating-point version. Comparing the integer models with smaller floating-point models in Table~\ref{tab:coco_analysis}, I-LW-DETR-Small outperforms the FP32 LW-DETR-Tiny by 1.1 mAP while being $2.7\times$ smaller and requiring $8.7\times$ fewer BOPs. I-LW-DETR-Medium outperforms it by 4.5 mAP ($1.5\times$ smaller, $3.8\times$ fewer BOPs), and comes within 0.9 mAP of the FP32 LW-DETR-Small while being $1.8\times$ smaller and requiring $5.6\times$ fewer BOPs. For a given memory and compute budget, a larger integer model is thus more accurate than a smaller floating-point one. %Moreover, on integer-only accelerators without floating-point units, the FP32 models cannot be executed natively at all. Model size and BOPs are, however, hardware-independent proxies. Measuring latency requires compiling the integer operators for a target accelerator, which is beyond the scope of this work (Section~\ref{ssec:mappability}).
}

\begin{figure}[!htbp]
    \centering
    \includegraphics[width=\linewidth]{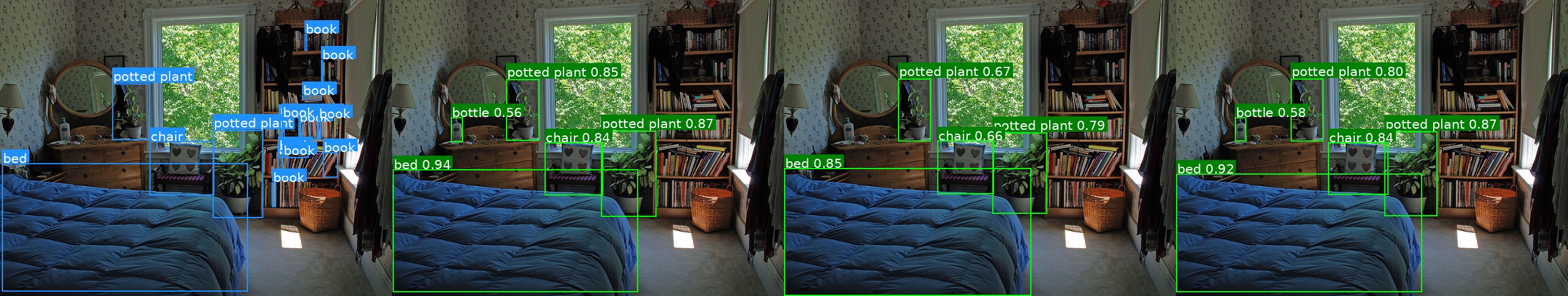}\\[-0.1pt]
    \includegraphics[width=\linewidth]{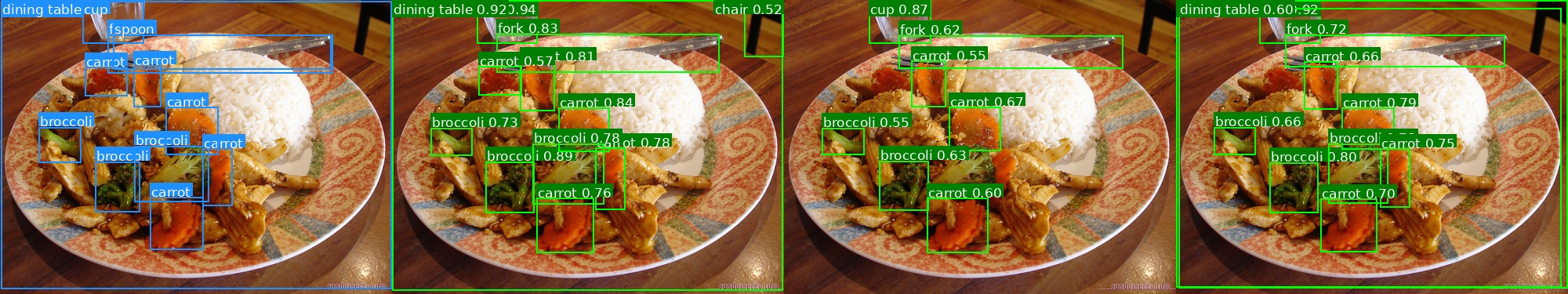}\\[-0.1pt]
    \includegraphics[width=\linewidth]{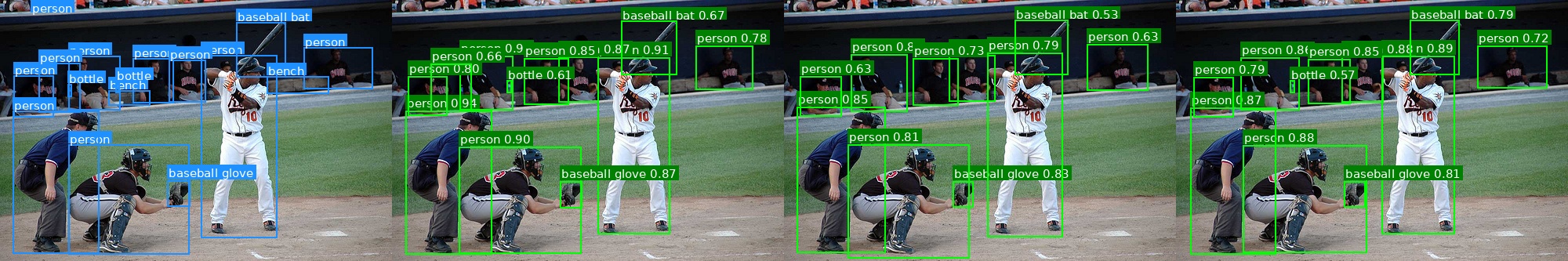}
    \caption{Qualitative detection results on the COCO dataset using Small backbone. From left to right: ground truth, QR-LW-DETR, PTQ I-LW-DETR, and QAT I-LW-DETR.}
    \label{fig:coco_stacked_results}
\end{figure}

Overall, these results demonstrate that the proposed quantization pipeline effectively transforms LW-DETR into fully integer-only models while maintaining a favorable trade-off between accuracy and deployment efficiency across different model scales. Figure~\ref{fig:coco_stacked_results} provides qualitative examples illustrating
the effect of PTQ and QAT on the detection outputs.

% ============================ VISDRONE ============================
\begin{table}[ht]
\centering
\footnotesize
\scriptsize
\setlength{\tabcolsep}{3pt}
\renewcommand{\arraystretch}{1.15}
\caption{Transfer to aerial imagery: I-LW-DETR on VisDrone.}
\label{tab:visdrone_results}
\begin{tabular}{llcccccccc}
\toprule
\textbf{Model} & \textbf{Setting} & \textbf{Prec.} & \textbf{Int-only} & \textbf{Size(MB)} & \textbf{BOPs(T)} & \textbf{mAP} & \textbf{AP\textsubscript{S}} & \textbf{AP\textsubscript{M}} & \textbf{AP\textsubscript{L}} \\
\midrule

 & LW-DETR & FP32   & \xmark & 44.89 & 11.48 & 15.7 & 6.9 & 24.6 & 40.5 \\
 & QR-LW-DETR       & FP32   & \xmark & 45.39 & 11.49 & 15.0 & 6.6 & 23.5 & 42.2 \\
\rowcolor{cyan!10}
 & PTQ      & 8/8/16 & \cmark & 12.57 & 0.92  & 7.8  & 2.3 & 13.4 & 27.3 \\
\rowcolor{cyan!22}
\multirow{-4}{*}{Tiny}
 & QAT      & 8/8/16 & \cmark & 12.57 & 0.92  & 10.9 & 4.1 & 17.6 & 31.9 \\

\midrule

 & LW-DETR & FP32   & \xmark & 54.44 & 17.0 & 18.2 & 8.1 & 28.6 & 48.1 \\
 & QR-LW-DETR       & FP32   & \xmark & 54.95 & 17.04 & 17.2 & 7.7 & 27.2 & 47.3 \\
\rowcolor{cyan!10}
 & PTQ      & 8/8/16 & \cmark & 16.93 & 1.36  & 13.3 & 5.5 & 21.5 & 40.8 \\
\rowcolor{cyan!22}
\multirow{-4}{*}{Small}
 & QAT      & 8/8/16 & \cmark & 16.93 & 1.36  & 15.6 & 6.6 & 25.0 & 43.3 \\

\midrule

 & LW-DETR & FP32   & \xmark & 106.63 & 43.78 & 19.3 & 8.9 & 30 & 49.3 \\
 & QR-LW-DETR      & FP32   & \xmark & 107.14 & 43.82 & 19 & 8.8 & 29.6 & 49.5 \\
\rowcolor{cyan!10}
 & PTQ      & 8/8/16 & \cmark & 30.16 & 3.07  & 11.4 & 4.2 & 19.6 & 37.1 \\
\rowcolor{cyan!22}
\multirow{-4}{*}{Medium}
 & QAT      & 8/8/16 & \cmark & 30.16 & 3.07  & 15 & 6.1 & 24.2 & 44.4 \\

\bottomrule
\end{tabular}
\end{table}

%\subsection{Qualitative Detection Results}

\label{sec:vis)}

\begin{figure}[!htbp]
    \centering
    \includegraphics[width=\linewidth]{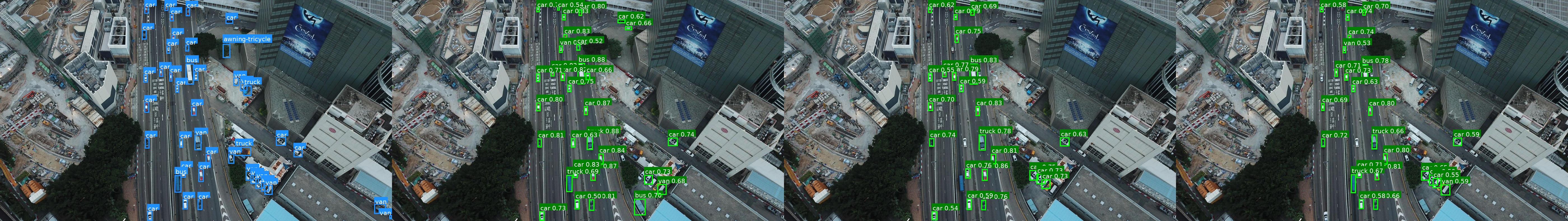}\\[-0.1pt]
    \includegraphics[width=\linewidth]{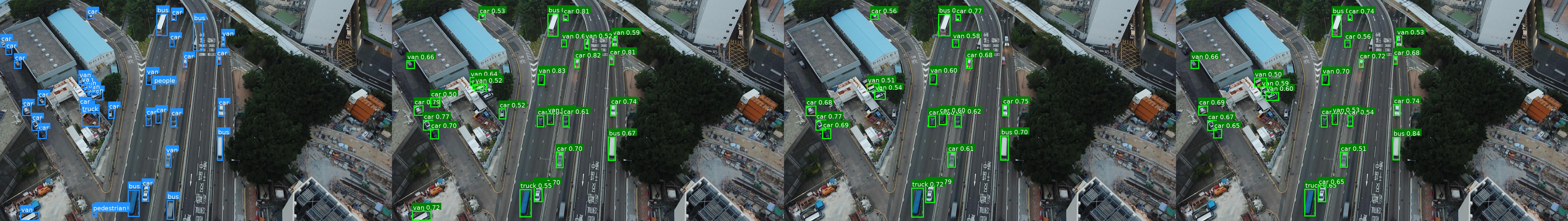}\\[-0.1pt]
    \includegraphics[width=\linewidth]{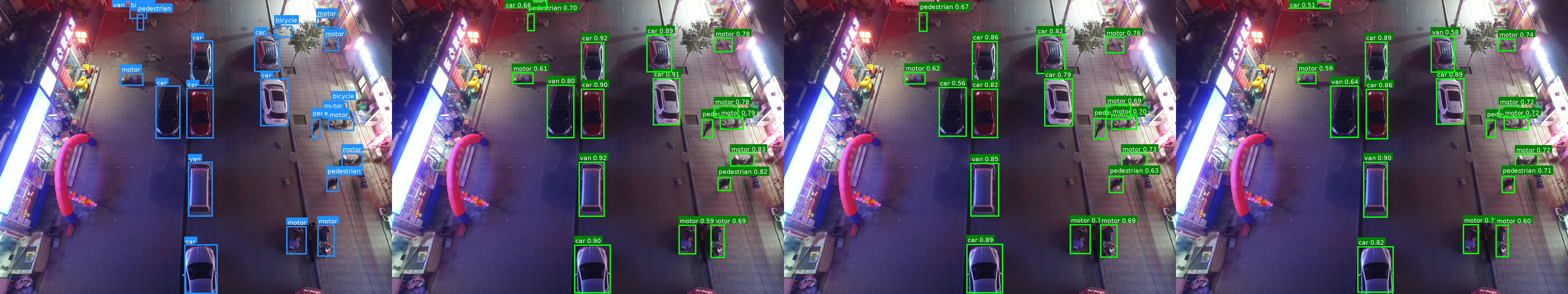}
    \caption{Qualitative detection results on the VisDrone dataset after fine-tuning, using Small backbone. From left to right: ground truth, QR-LW-DETR, PTQ I-LW-DETR, and QAT I-LW-DETR.}
    \label{fig:visdrone_stacked_results}
\end{figure}

Table~\ref{tab:visdrone_results} evaluates the transferability of the proposed quantization pipeline on VisDrone. Replacing the floating-point operators with their quantization-ready counterparts results in only a minor accuracy degradation before quantization, indicating that the architectural modifications remain effective beyond the COCO benchmark. Applying PTQ preserves the same reduction in model size and computational cost as on COCO but results in a substantially larger accuracy degradation, reflecting the higher sensitivity of aerial small-object detection to integer quantization. This degradation is particularly pronounced for small objects, where AP\textsubscript{S} exhibits the largest drop from the QR baseline to PTQ across all model variants. QAT consistently recovers a significant portion of the lost accuracy without increasing the deployment cost, although the recovery remains more limited for small objects than for medium and large ones. Figure~\ref{fig:visdrone_stacked_results} provides qualitative examples illustrating that, in these dense scenes, small and closely spaced objects remain more frequently missed or confused. One possible contributing factor is the replacement of bilinear interpolation with nearest-neighbor sampling in deformable attention, which may reduce the localization precision required for very small objects. 
Overall, these results demonstrate that the proposed integer-only pipeline generalizes well to aerial imagery, while highlighting that aggressive quantization remains considerably more challenging for small-object detection than for natural-image benchmarks such as COCO.

\subsection{Comparison with Existing Quantized DETR Models}
% Table~\ref{tab:coco_comparison} compares I-LW-DETR with existing quantized DETR methods on COCO. Since these methods use different detector architectures and backbones, the comparison should be interpreted as contextual rather than strictly one-to-one. Among them, EIQ-DETR is the closest reference because it is the only previous fully integer-only DETR method reported in the table.

\rev{Table~\ref{tab:coco_comparison} places I-LW-DETR among existing quantized DETR methods on COCO. These methods differ in detector architecture, backbone, attention type, and quantization setting. The table should therefore be interpreted as a system-level comparison: differences in model size, BOPs, and mAP reflect the combined effect of architecture and quantization rather than the quantization scheme alone. EIQ-DETR is the closest reference, as it is the only other fully integer-only DETR in the comparison.}

\begin{table*}[ht]
\centering
\scriptsize
\setlength{\tabcolsep}{4pt}
\renewcommand{\arraystretch}{1.15}
%OLD caption (R1.3): \caption{Comparison with existing quantized DETR methods on COCO.}
\caption{\rev{System-level comparison with existing quantized DETR methods on COCO. Methods differ in detector architecture, backbone, and attention mechanisms; therefore, performance differences reflect both architectural and quantization choices. Attention types: G = global, W = window-based, D = deformable, H = hybrid.}}
%\caption{\michal{System-level comparison with existing quantized DETR methods on COCO. The methods differ in detector architecture, backbone and attention type, so differences reflect the combined effect of architecture and quantization. %operator-level comparisons on a common detector are given in Tables~\ref{tab:gelu_map_comparison} and~\ref{tab:shiftmax_ablation}.Attention: G = global, W = window, D = deformable, H = hybrid}}
\label{tab:coco_comparison}
%OLD table (R1.3): columns Model | Setting | Precision | Int-only | Size(MB) | BOPs(T) | mAP; our values in bold. New columns in orange. FP32 mAP and $\Delta$ columns removed.
\setlength{\tabcolsep}{3pt}
\resizebox{\textwidth}{!}{%
\begin{tabular}{lllcccccc}
\toprule
\textbf{Model} & \textbf{Detector / Backbone} & \textbf{Attn.} & \textbf{Setting} & \textbf{Prec.} & \textbf{Int-only} & \shortstack{\textbf{Size}\\\textbf{(MB)}} %\textbf{Size(MB)}& \textbf{BOPs(T)} 
& \shortstack{\textbf{BOPs}\\\textbf{(T)}} & \textbf{mAP} \\
\midrule
QwT              & DETR / R50   & G & PTQ & 8/8 & \xmark & 58.4 & -- & 41.7 \\
IGQ-ViT          & DETR / R50   & G & PTQ & 6/6 & \xmark &    & -- & 41.3 \\
Reg-PTQ          & DETR / R50   & G & PTQ & 4/4 & \xmark &    & -- & 37.4 \\
Q-DETR           & DETR / R50                & G   & QAT & 4/4/8 & \xmark & 19.9 & -- & 39.4 \\
AQ-DETR          & Def.\ DETR / R50          & D   & QAT & 4/4/4 & \xmark & 24.9 & -- & 44.1 \\
QRT-DETR         & RT-DETR / R18             & H/D & PTQ & 8/8/8 & \xmark & 20.25 & -- & 46.3 \\
EIQ-DETR         & DETR / Swin-T             & W/G & QAT & 8/8/8 & \cmark & 43.49 & 5.90 & 44.1 \\
\midrule
\rowcolor{cyan!12}
I-LW-DETR  & LW-DETR / Tiny             & W/G/D & QAT & 8/8/16 & \cmark & 12.85 & 0.89 & 37.9 \\
\rowcolor{cyan!12}
%I-LW-DETR-Small  
& LW-DETR / Small             & W/G/D & QAT & 8/8/16 & \cmark & 17.21 & 1.32 & 43.7 \\
\rowcolor{cyan!12}
%I-LW-DETR-Medium
& LW-DETR / Medium             & W/G/D & QAT & 8/8/16 & \cmark & 30.44 & 3.04 & 47.1 \\
\bottomrule
\end{tabular}}
\end{table*}

\rev{Among the fully integer-only methods, I-LW-DETR-Medium achieves 47.1 mAP with 30.44 MB and 3.04 TBOPs, compared with 44.1 mAP, 43.49 MB, and 5.90 TBOPs for EIQ-DETR. Thus, even the largest I-LW-DETR variant has a smaller model footprint and lower reported BOPs while achieving higher detection accuracy. I-LW-DETR uses an 8/8/16 configuration, whereas EIQ-DETR reports 8/8/8 precision and does not address deformable attention. The remaining baselines are not fully integer-only and use different precision settings; they are therefore included mainly to provide context on the accuracy of existing quantized DETR approaches rather than for direct computational-cost comparison.}

\section{Ablation studies}
\label{sec:ablation}

\subsection{Component-wise quantization}

To assess whether existing integer operators generalize to DETR-based object detection, we individually replace each floating-point operation by its I-ViT integer counterpart while keeping the remainder of the network in floating-point precision. Table~\ref{tab:ablation_study} shows that several integer operators originally proposed for ViT image classification do not generalize well to LW-DETR.

\begin{table}[!h]
\centering
\scriptsize
\setlength{\tabcolsep}{4pt}
\renewcommand{\arraystretch}{1.15}
\caption{Component-wise PTQ ablation study based on the I-ViT quantization method. Each row quantizes one operation while keeping the rest of the model in floating-point precision. The symbol \(^{\star}\) indicates sensitive components.}
\label{tab:ablation_study}
\resizebox{\textwidth}{!}{%
\begin{tabular}{lllccc}
\toprule
\textbf{Component} & \textbf{Operation} & \textbf{Quantized module} & \textbf{Tiny} & \textbf{Small} & \textbf{Medium} \\
\midrule

QR-LW-DETR FP32 & -- & -- & 42.0 & 46.4 & 50.8 \\

\midrule
\multirow{6}{*}{ViT encoder}
& Softmax\(^{\star}\) & Shiftmax\(^{\star}\)  & 6.3  & 42.3 & 44.0 \\
& GELU\(^{\star}\)    & ShiftGELU\(^{\star}\) & 0.0  & 8.2  & 2.1  \\
& Linear              & QuantLinear           & 41.8 & 46.0 & 50.6 \\
& Conv2d              & QuantConv2d           & 41.9 & 46.4 & 50.8 \\
& MatMul              & QuantMatMul           & 41.9 & 46.2 & 50.7 \\
& LayerNorm           & I-LayerNorm           & 42.0 & 46.4 & 50.8 \\

\midrule
\multirow{3}{*}{Projector}
& Conv2d\(^{\star}\) & QuantConv2d\(^{\star}\) & 41.5 & 41.7 & 37.6 \\
& SiLU               & ShiftSiLU               & 41.3 & 45.1 & 49.5 \\
& LayerNorm          & I-LayerNorm             & 41.3 & 45.5 & 49.0 \\

\midrule
\multirow{5}{*}{Transformer decoder}
& Linear      & QuantLinear & 41.2 & 45.2 & 48.6 \\
& MatMul      & QuantMatMul & 41.7 & 46.4 & 50.6 \\
& Softmax     & Shiftmax    & 41.7 & 46.3 & 50.4 \\
& LayerNorm   & I-LayerNorm & 41.3 & 46.0 & 49.4 \\
& Exponential & ShiftExp    & 41.8 & 46.3 & 50.6 \\

\midrule
\multirow{2}{*}{Detection head}
& Linear      & QuantLinear & 41.1 & 46.2 & 50.3 \\
& Exponential & ShiftExp    & 39.0 & 44.6 & 48.5 \\

\bottomrule
\end{tabular}%
}
\end{table}

Most linear transformations, matrix multiplications, and normalization layers remain highly robust, exhibiting less than 1 mAP degradation compared with the FP32 baseline across all three model variants. In contrast, the I-ViT ShiftGELU causes severe accuracy degradation, reducing the detection performance to 0.0, 8.2, and 2.1 mAP for the Tiny, Small, and Medium models, respectively. Similarly, the Shiftmax reduces the Tiny model from 42.0 to 6.3 mAP, while quantizing the projector Conv2d lowers the Medium model from 50.8 to 37.6 mAP. These observations indicate that the original I-ViT quantization scheme, developed for ViT image classification, cannot be directly transferred to DETR-based object detection without architectural adaptations, thereby motivating the proposed projector convolution, SD-ShiftGELU, and Constrained Shiftmax presented in the following subsections.

\subsection{Cumulative contribution analysis}

While the previous experiment isolates the effect of each individual operation, Table~\ref{tab:cumulative_ablation} evaluates the incremental contribution of the proposed components under a fully integer-only setting. Starting from a fully quantized LW-DETR using the original I-ViT quantization scheme, the proposed components are introduced sequentially, with each row accumulating the modifications introduced in the previous rows.

\begin{table}[!h]
\centering
\scriptsize
\setlength{\tabcolsep}{4pt}
\renewcommand{\arraystretch}{1.15}
\caption{Incremental PTQ ablation. Starting from the naive I-ViT quantization scheme, the proposed components are introduced sequentially. Each row accumulates the modifications from the previous rows while all remaining operations remain quantized.}
\label{tab:cumulative_ablation}
\begin{tabular}{lccc}
\toprule
\textbf{Configuration} & \textbf{Tiny} & \textbf{Small} & \textbf{Medium} \\
\midrule
QR-LW-DETR FP32 upper bound & 42.0 & 46.4 & 50.8 \\
\midrule
Naive I-ViT, all quantized & 0.0 & 1.3 & 3.6 \\
\hspace{1em} + SD-ShiftGELU & 3.9 & 31.2 & 24.3 \\
\hspace{2em} + Constrained Shiftmax & 36.0 & 36.7 & 32.2 \\
\hspace{3em} + Split Conv2d & 36.0 & 41.6 & 44.0 \\
\bottomrule
\end{tabular}
\end{table}

The first row highlights the limitation of directly applying the I-ViT quantization scheme to QR-LW-DETR. Under PTQ, the detection accuracy drops dramatically from 42.0, 46.4, and 50.8 mAP to only 0.0, 1.3, and 3.6 mAP for the Tiny, Small, and Medium models, respectively. Replacing the original ShiftGELU with the proposed SD-ShiftGELU substantially restores the performance of the Small and Medium models, increasing their accuracy from 1.3 to 31.2 mAP and from 3.6 to 24.3 mAP, respectively. In contrast, only a limited improvement is observed for the Tiny model (0.0 to 3.9 mAP), indicating that additional quantization bottlenecks remain dominant in this configuration. Introducing the proposed Constrained Shiftmax provides another substantial performance gain, particularly for the Tiny model, highlighting the importance of accurate integer Softmax normalization. Finally, the proposed split convolution delivers further improvements, especially for the Small and Medium models, yielding the final PTQ performance of 36.0, 41.6, and 44 mAP for the Tiny, Small, and Medium variants, respectively, corresponding to the complete I-LW-DETR PTQ model reported in Section~\ref{sec:results}. %\michal{Each component therefore addresses a distinct bottleneck and yields a substantial gain at its stage of the sequence. The only exception is the split convolution for the Tiny model, whose projector receives fewer encoder outputs (Table~\ref{tab:lwdetr_variants}). These bottlenecks lie in the GELU, the attention normalization and the multi-scale projector, the last of which is absent from the vanilla DETR architecture considered in previous fully integer-only work~\cite{eiqdetr}.}
\rev{These results highlight three distinct quantization bottlenecks associated with GELU activation, attention normalization, and multi-scale feature projection. Their relative impact varies across model sizes, with split convolution providing limited benefits for Tiny, whose projector processes fewer encoder outputs. Notably, the multi-scale projector introduces a quantization challenge absent from the vanilla DETR architecture addressed by previous fully integer-only work~\cite{eiqdetr}.}

\subsection{Split Convolution}
\label{sec:ablation-splitconv}

Table~\ref{tab:split_conv_projector} isolates the proposed split convolution in the projector. The benefit is negligible for the Tiny model but becomes increasingly significant for the larger variants, with the Medium model improving from 37.6 to 49.1 mAP. This trend suggests that projector quantization becomes increasingly sensitive to scale mismatch as the feature dimensionality grows.

\begin{table}[!ht]
\centering
\scriptsize
\setlength{\tabcolsep}{4pt}
\renewcommand{\arraystretch}{1.15}
\caption{Effect of scale-aware split convolution in the projector.}
\label{tab:split_conv_projector}
\begin{tabular}{lcc}
\toprule
\textbf{Model} 
& \textbf{Concat $\rightarrow$ Conv2d} 
& \textbf{Split Conv2d} \\
\midrule
Tiny   & 41.5 & 41.5 \\
Small  & 41.7 & \textbf{45.9} \\
Medium & 37.6 & \textbf{49.1} \\
\bottomrule
\end{tabular}
\end{table}

\begin{figure}[!htbp]
    \centering
    \subfloat[\footnotesize Activation distributions before feature concatenation]{
        \includegraphics[width=0.95\linewidth]{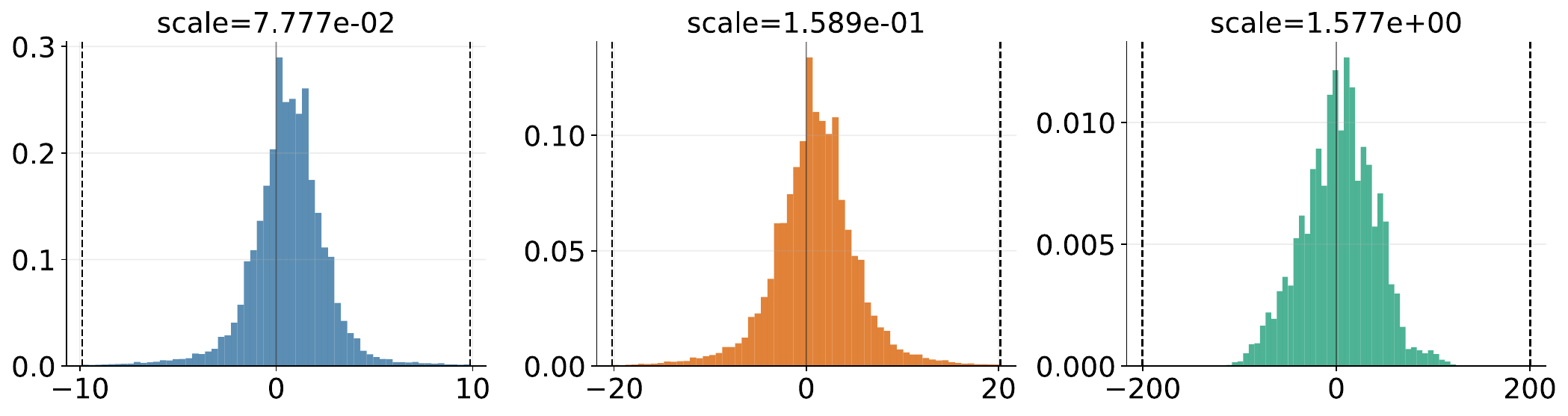}
        \label{fig:hist_shared_scale_conv}
    }

    \vspace{0.5em}

    \subfloat[\footnotesize Activation distributions after feature concatenation]{
        \includegraphics[width=0.95\linewidth]{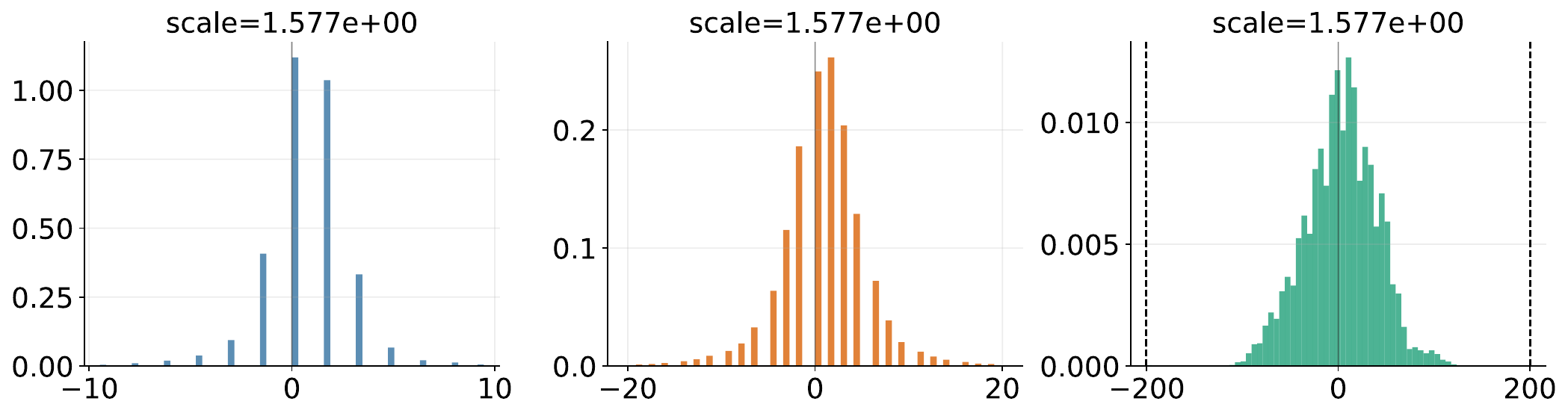}
        \label{fig:hist_split_scale_conv}
    }

    \caption{Activation distributions before and after feature concatenation.}
    \label{fig:hist_projector_split_conv}
\end{figure}

Figure~\ref{fig:hist_projector_split_conv} visualizes the cause. After feature concatenation, all channels are quantized using a single activation scale determined by the largest-range feature group. Consequently, feature groups with smaller dynamic ranges lose quantization resolution. The proposed split convolution avoids this issue by preserving an independent activation scale for each feature group.

\subsection{GELU Integer Approximation Methods}
As shown in Table \ref{tab:ablation_study}, GELU is one of the most quantization-sensitive operations in QR-LW-DETR. \rev{To further investigate this behavior, Table~\ref{tab:gelu_map_comparison} compares four integer GELU implementations: the original ShiftGELU from I-ViT~\cite{ivit}, the $\lambda$-ShiftGELU from I-Segmenter~\cite{isegmenter}, the integer GELU approximation used in EIQ-DETR~\cite{eiqdetr}, and the proposed
SD-ShiftGELU. All methods are evaluated under identical conditions, using the same pretrained model, calibration data, activation ranges, and bit-width configuration, with only the GELU approximation being changed.}
%Table~\ref{tab:gelu_map_comparison} therefore compares the proposed SD-ShiftGELU with three existing integer GELU approximations from I-ViT, I-Segmenter, and EIQ-DETR. SD-ShiftGELU consistently achieves the highest detection accuracy across all model sizes. While the original I-ViT approximation causes catastrophic performance degradation, SD-ShiftGELU consistently outperforms the more recent I-Segmenter and EIQ-DETR approximations, suggesting that the proposed sign-dependent formulation more faithfully reproduces the floating-point GELU under integer arithmetic.

\begin{table}[!h]
\centering
\scriptsize
\setlength{\tabcolsep}{4pt}
\renewcommand{\arraystretch}{1.15}
\caption{Detection performance when only GELU is quantized.}
\label{tab:gelu_map_comparison}
\begin{tabular}{lcccc}
\toprule
\textbf{Model size} 
& \multicolumn{4}{c}{\textbf{GELU approximation}} \\
\cmidrule(lr){2-5}
& \textbf{I-ViT} 
& \textbf{I-Segmenter} 
& \textbf{EIQ-DETR} 
& \textbf{SD-ShiftGELU} \\
\midrule
Tiny   & 0.0  & 37.0 & 41.0 & \textbf{41.3} \\
Small  & 8.2  & 42.0 & 45.0 & \textbf{45.6} \\
Medium & 2.1  & 44.5 & 49.3 & \textbf{50.1} \\
\bottomrule
\end{tabular}
\end{table}

\rev{The original I-ViT ShiftGELU causes severe performance degradation, reaching only 0.0, 8.2, and 2.1 mAP for the Tiny, Small, and Medium models, respectively. In contrast, I-Segmenter and EIQ-DETR
substantially recover the detection accuracy, while SD-ShiftGELU achieves
the highest performance across all three model sizes. To better understand
the origin of these differences, we further analyze the numerical behavior
of the four GELU approximations.}

\begin{figure}[H]
    \centering
    \begin{subfigure}[t]{0.48\linewidth}
        \centering
        \includegraphics[width=\linewidth]{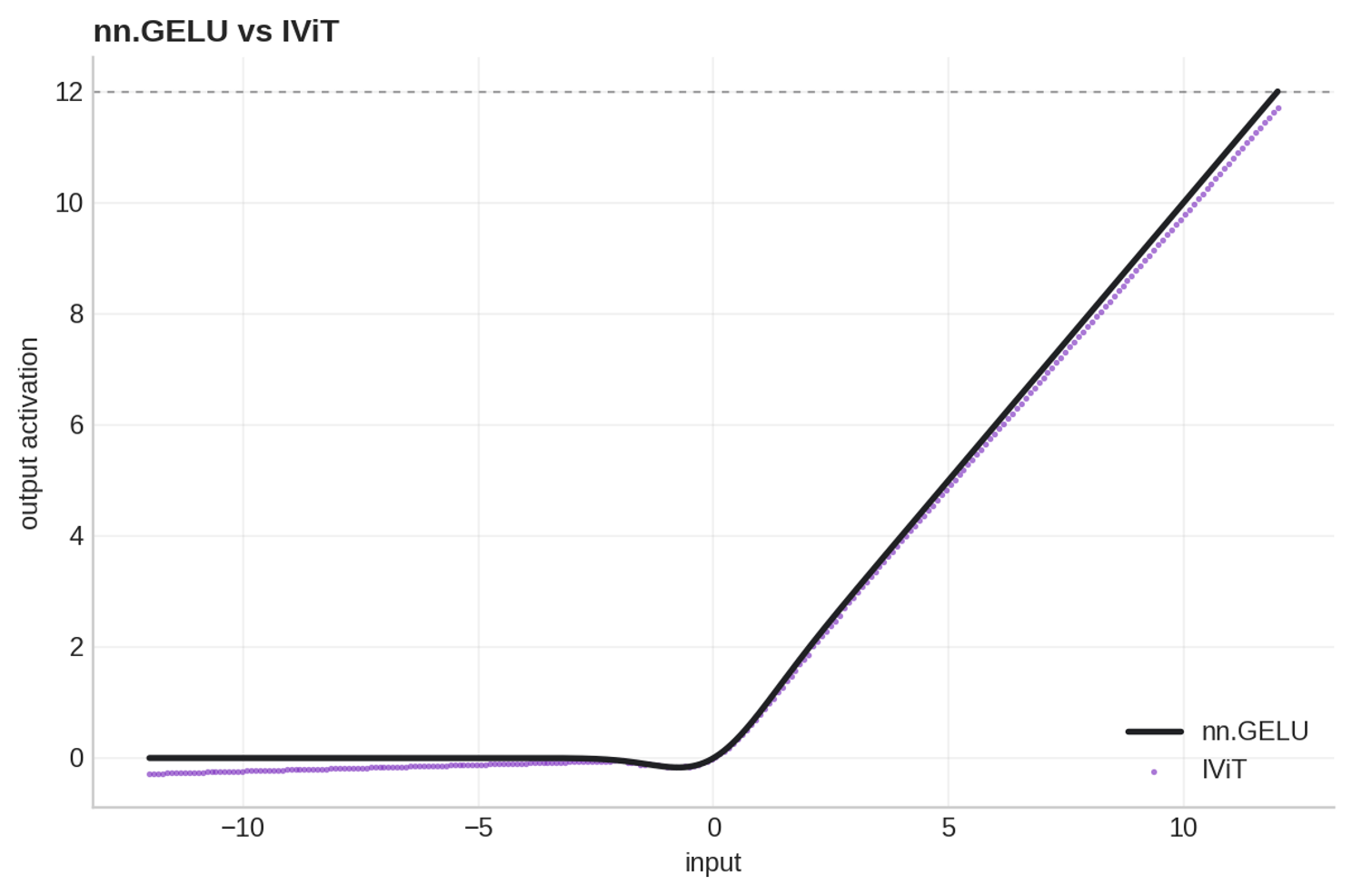}
        \caption{I-ViT}
    \end{subfigure}
    \hfill
    \begin{subfigure}[t]{0.48\linewidth}
        \centering
        \includegraphics[width=\linewidth]{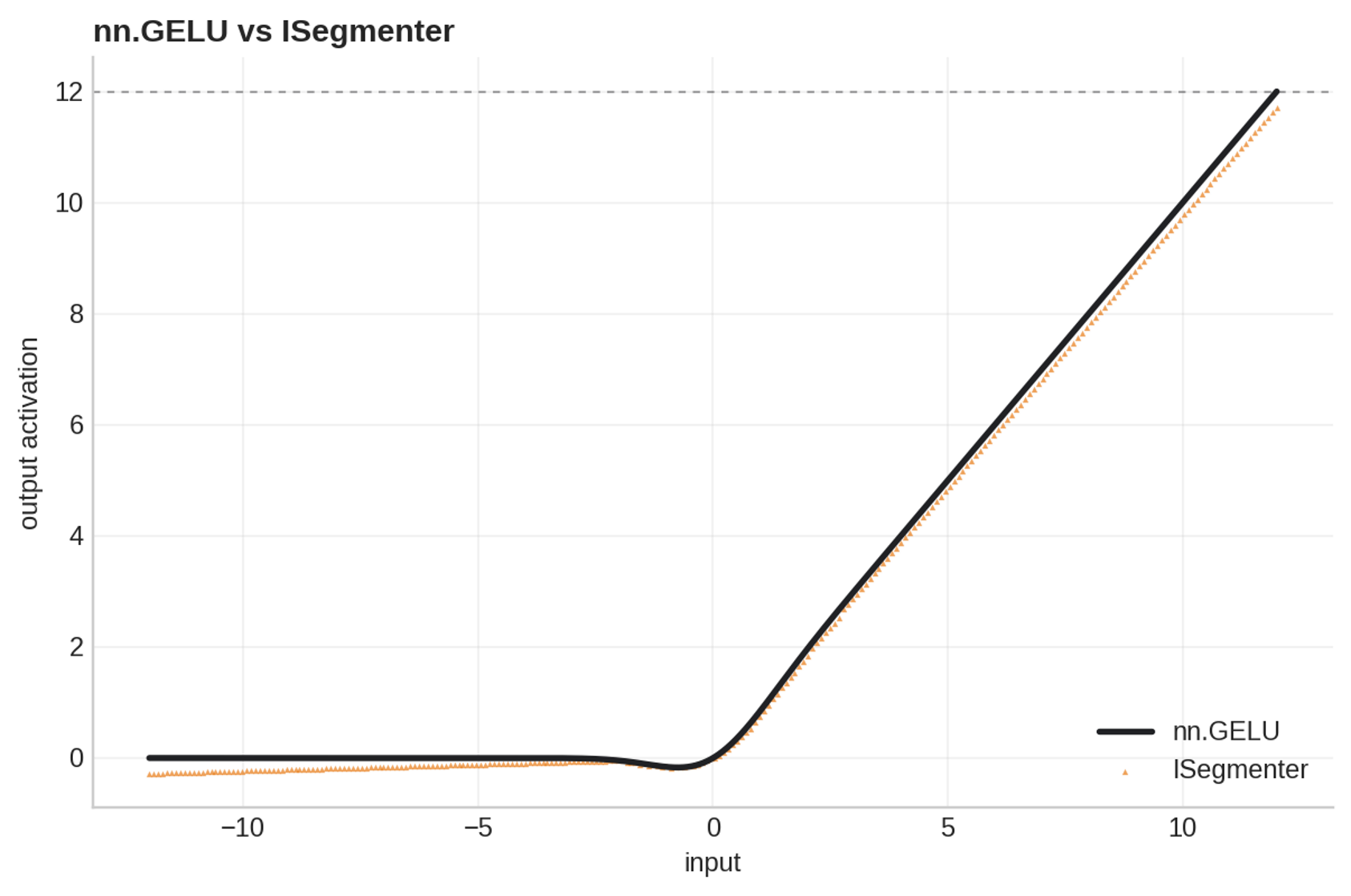}
        \caption{I-Segmenter}
    \end{subfigure}

    \vspace{0.5em}

    \begin{subfigure}[t]{0.48\linewidth}
        \centering
        \includegraphics[width=\linewidth]{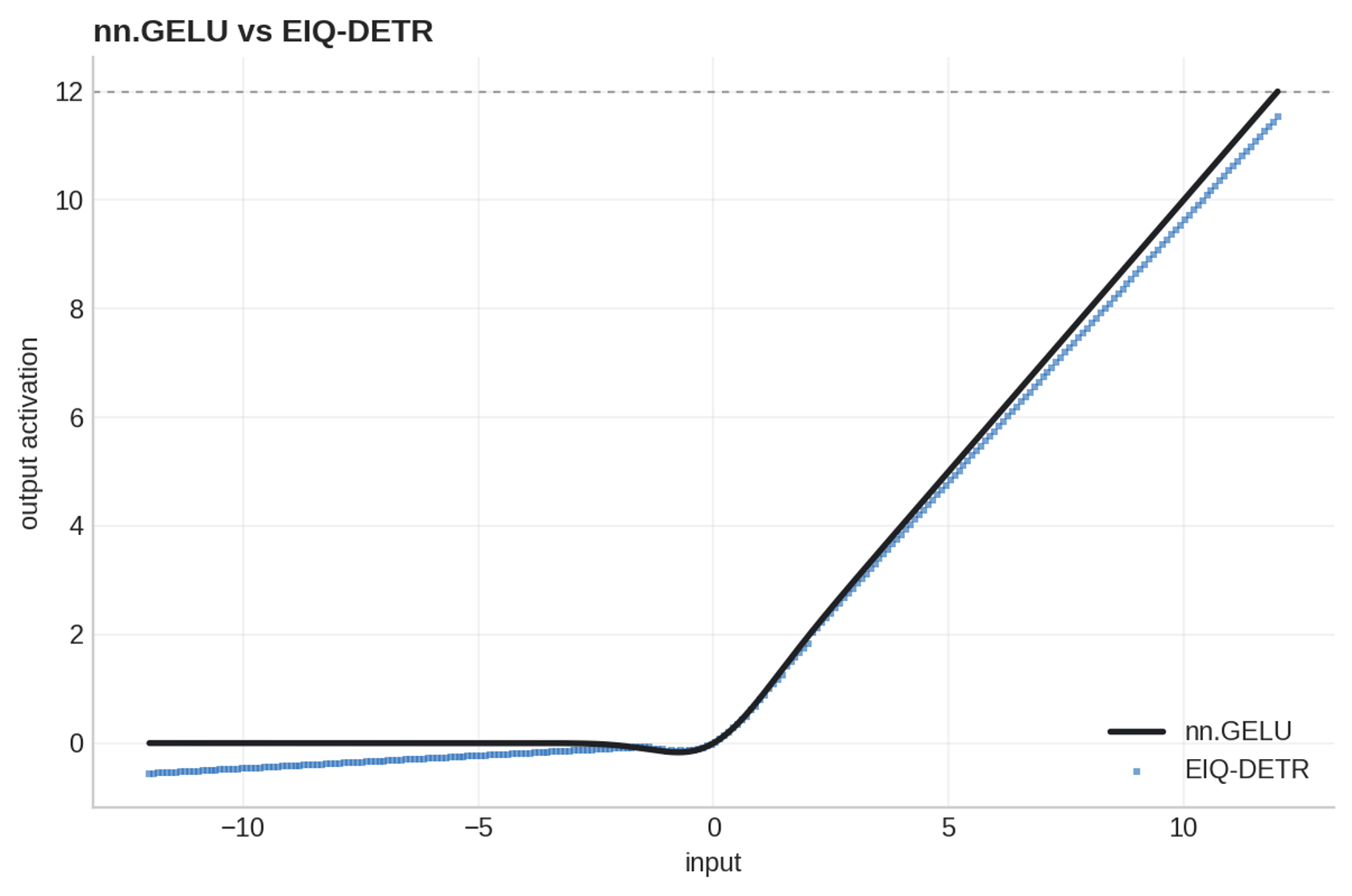}
        \caption{EIQ-DETR}
    \end{subfigure}
    \hfill
    \begin{subfigure}[t]{0.48\linewidth}
        \centering
        \includegraphics[width=\linewidth]{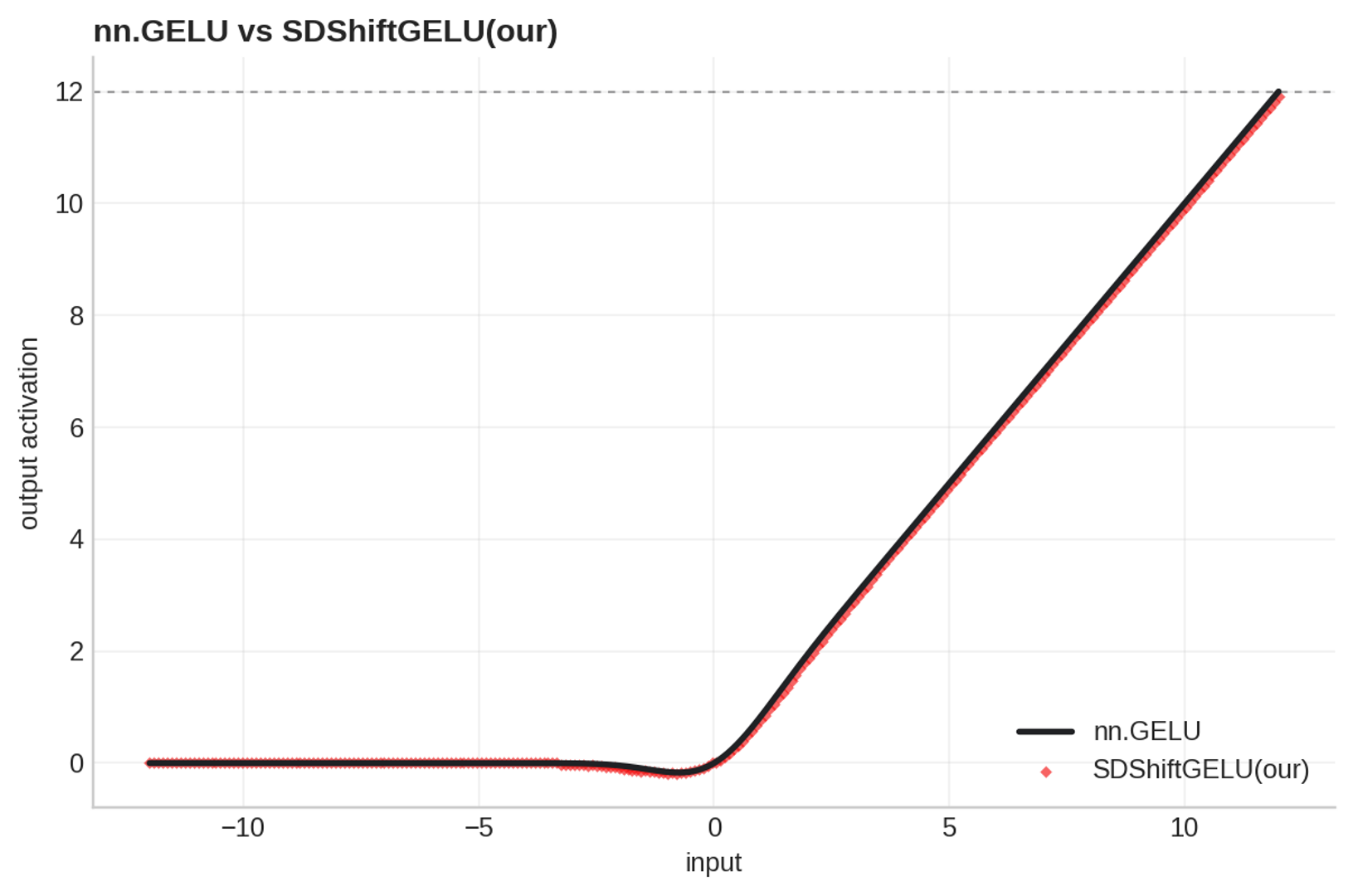}
        \caption{SDShiftGELU}
    \end{subfigure}

    \caption{Comparison of integer GELU approximations with floating-point
    \texttt{nn.GELU} over a controlled input range.}
    \label{fig:gelu_activation_curves}
\end{figure}

\rev{To compare the approximation behavior over a controlled input range,
Figure~\ref{fig:gelu_activation_curves} shows the four integer GELU
approximations against floating-point \texttt{nn.GELU}. The curves are
generated using 255 uniformly spaced input values between $-12$ and $12$.
I-ViT ShiftGELU exhibits the largest deviation, particularly for negative
inputs, whereas SD-ShiftGELU follows the floating-point reference more
closely over the considered range.}

\begin{figure}[H]
    \centering
    \includegraphics[width=0.85\linewidth]{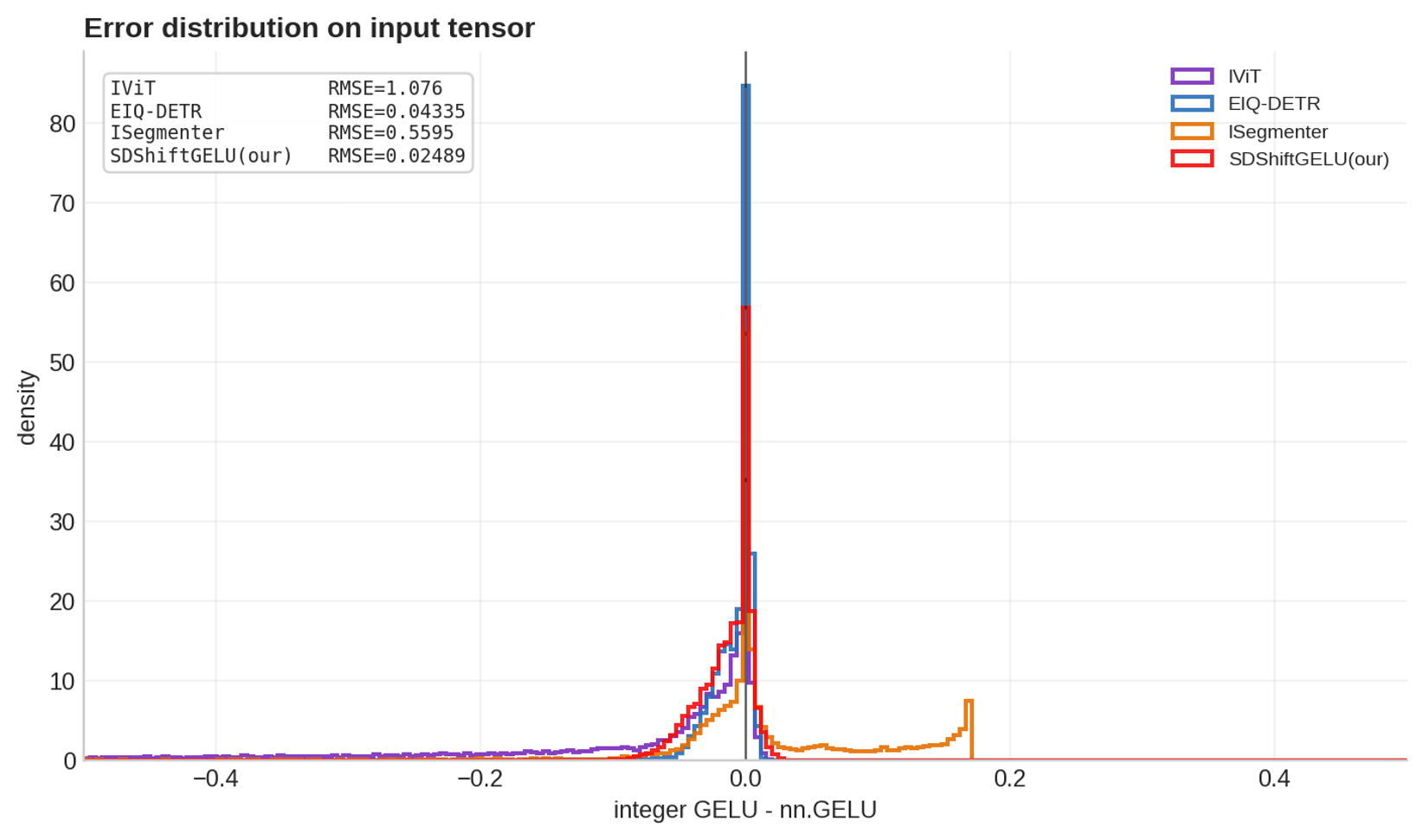}
    \caption{Distribution of the GELU approximation error relative to
    floating-point \texttt{nn.GELU}, evaluated on actual activations
    collected from the first ViT encoder block.}
    \label{fig:gelu_error_distribution}
\end{figure}

\rev{We further evaluate the approximation error on actual GELU input
activations collected from the first ViT encoder block. Figure~\ref{fig:gelu_error_distribution} shows the error distributions
relative to floating-point \texttt{nn.GELU}. I-ViT exhibits the largest error, with an RMSE of 1.076, compared with 0.5595 for I-Segmenter,
0.04335 for EIQ-DETR, and 0.02489 for SD-ShiftGELU. Since all methods are evaluated using the same calibration data, activation ranges, and bit-width configuration, these results indicate
that the severe degradation observed with I-ViT is associated with its GELU approximation rather than differences in the quantization setup.}

\subsection{Constrained Shiftmax}

\begin{table}[ht]
\centering
\scriptsize
\setlength{\tabcolsep}{4pt}
\renewcommand{\arraystretch}{1.15}
\caption{Effect of the proposed Constrained Shiftmax on detection performance when only Softmax is quantized.}
\label{tab:shiftmax_ablation}
\begin{tabular}{lccc}
\hline
\textbf{Method} & \textbf{Tiny} & \textbf{Small} & \textbf{Medium} \\
\hline
I-ViT Shiftmax & 5.5 & 42.2 & 44.4 \\
Constrained Shiftmax & \textbf{41.8} & \textbf{45.9} & \textbf{50.3} \\
\hline
\end{tabular}
\end{table}

Table~\ref{tab:shiftmax_ablation} isolates the proposed Constrained Shiftmax by quantizing only the Softmax operation. The original I-ViT Shiftmax causes severe performance degradation, particularly for the Tiny model, where the detection accuracy drops to 5.5 mAP. This behavior occurs because the accumulated exponential sum may exceed the fixed integer precision \(M\) in some attention blocks, causing the normalization factor to collapse toward zero. The proposed Constrained Shiftmax restores high detection accuracy across all model sizes by calibrating the denominator shift \(s_d\) for each attention block. These results demonstrate that adapting the normalization range is essential for accurate integer Softmax in LW-DETR.
  % version finale

%\section{Discussion: Hardware Compatibility}
%\label{sec:mappability}

\section{\rev{Discussion}}
\label{sec:discussion}

\subsection{\rev{Main Findings and Implications}}

\rev{The experiments show that directly transferring existing integer transformer operators from prior vision tasks to a lightweight deformable DETR can lead to severe accuracy degradation. The naive I-ViT-based quantization pipeline causes a substantial collapse in detection accuracy, whereas the proposed SD-ShiftGELU, Constrained Shiftmax, and scale-preserving split convolution progressively restore performance. This indicates that the main challenge is not the quantization of linear layers, which remain comparatively robust, but the interaction between detector-specific activation distributions, nonlinear operators, and multi-scale feature processing.}

\rev{Compared with previous DETR quantization approaches, I-LW-DETR targets a distinct deployment setting. Q-DETR, Reg-PTQ, IGQ-ViT, and QwT report quantization results on standard DETR-based architectures, while AQ-DETR and QRT-DETR also address detectors with deformable attention. However, these approaches focus primarily on low-bit weight and activation quantization and do not provide fully integer-only inference over the complete DETR computational graph. EIQ-DETR achieves fully integer-only inference, but does not address deformable attention. In contrast, I-LW-DETR combines fully integer-only inference with a lightweight deformable DETR architecture.}

\subsection{\rev{Hardware Compatibility}}
\label{ssec:mappability}

%A fully integer-only computational graph is necessary for deployment on integer edge accelerators, but it is not sufficient. The operators composing the graph must also correspond to primitives commonly exposed by the target hardware. To assess this compatibility, we compare the operators required by I-LW-DETR with the published integer operator set of the Arm Ethos-U85 \footnote{ \url{https://developer.arm.com/documentation/102684/0000}.} and the corresponding Vela compiler documentation \footnote{ \url{https://gitlab.arm.com/artificial-intelligence/ethos-u/ethos-u-vela}.}, used here as a representative modern integer edge accelerator. 
\rev{A fully integer-only computational graph is necessary for deployment on integer edge accelerators, but it is not sufficient. The operators composing the graph must also correspond to primitives commonly exposed by the target hardware. To assess this compatibility, we compare the operators required by I-LW-DETR with the published integer operator set of the Arm Ethos-U85%
\footnote{\url{https://developer.arm.com/documentation/102684/0000}.}
and the corresponding Vela compiler documentation%
\footnote{\url{https://gitlab.arm.com/artificial-intelligence/ethos-u/ethos-u-vela}.}
Unlike FPGA platforms, where custom datapaths can be implemented for unsupported operations, the Ethos-U85 exposes a fixed and documented operator set, making it suitable for assessing compatibility with constrained integer edge accelerators.} This analysis
evaluates operator compatibility only; it does not constitute a deployment
study, and no compilation or on-device measurements are reported. % most I-LW-DETR operators map directly to standard integer primitives available on modern NPUs, including convolutions, fully connected layers, and matrix multiplication. 
%As summarized in Table~\ref{tab:mappability}, the I-LW-DETR that account for essentially all of the arithmetic operators map directly to standard integer primitives available on modern NPUs, including convolutions, fully connected layers, and matrix multiplication. 
\revMi{As summarized in Table~\ref{tab:mappability}, the I-LW-DETR operators that account for essentially all of the arithmetic, namely convolutions, fully connected layers and matrix multiplications, map directly to standard integer primitives available on modern NPUs.}

Transformer non-linearities remain the main compatibility challenge, as dedicated GELU and Softmax primitives are generally unavailable on integer accelerators. Consequently, their implementation relies only on standard integer arithmetic, while hardware-specific primitives, when available (e.g., \texttt{LOGISTIC} or \texttt{SOFTMAX} on the Ethos-U85), may serve as optional implementation alternatives. Some operations, including ShiftExp, I-LayerNorm, and deformable sampling, involve data-dependent computations or indexing. Although their mapping to hardware resources may depend on the compiler implementation, they remain entirely expressible using integer arithmetic and standard integer operations. % Consequently, they do not compromise the fully integer-only execution model.}

\begin{table}[t]
\centering
\scriptsize
\setlength{\tabcolsep}{4pt}
\renewcommand{\arraystretch}{1.15}
\caption{Representative correspondence between I-LW-DETR operators and integer primitives supported by the Arm Ethos-U85. The table illustrates operator compatibility only and does not imply compiler placement or hardware deployment.}
\label{tab:mappability}
\begin{tabular}{ll}
\hline
\textbf{I-LW-DETR operator} & \textbf{Representative Ethos-U85 primitive(s)} \\
\hline
Patch embedding / projections & \texttt{CONV\_2D}, \texttt{FULLY\_CONNECTED}, \texttt{ADD} \\
Attention & \texttt{MATMUL} \\
Scale-preserving split convolution & \texttt{CONV\_2D} $\times N$, \texttt{ADD} \\
Positional projection & \texttt{FULLY\_CONNECTED} \\
SD-ShiftGELU & Integer arithmetic / \texttt{LOGISTIC} (optional) \\
Calibrated Shiftmax & Integer arithmetic / \texttt{SOFTMAX} (optional) \\
I-LayerNorm & \texttt{MEAN}, integer arithmetic \\
Deformable sampling & \texttt{GATHER} \\
\hline
\end{tabular}
\end{table}

\subsection{\rev{Limitations}}

\rev{Several limitations remain. First, the present study evaluates computational cost and hardware compatibility but does not report on-device latency or energy measurements. Such measurements require a complete deployment of I-LW-DETR on a target accelerator, including compiler support or dedicated integer implementations for the proposed operators, and verification that the entire graph executes without floating-point fallback. Standard GPU fake-quantization or generic INT8 runtimes would not faithfully represent the proposed fully integer-only pipeline. Implementing and compiling these operators for a specific target, followed by end-to-end profiling, is therefore left for future work. Second, the experiments focus on the 8/8/16 quantization setting, and more aggressive low-bit configurations remain to be investigated. Third, the proposed approach is evaluated on the LW-DETR family, and its generalization to other deformable or efficient DETR architectures, such as Deformable DETR~\cite{deformabledetr}, RT-DETR/RT-DETRv2~\cite{rtdetr,rtdetrv2}, and Lite DETR~\cite{li2023litedetr}, remains to be investigated. Finally, the VisDrone experiments show that small-object detection is more sensitive to integer-only inference, potentially due in part to replacing bilinear interpolation with nearest-neighbor sampling in deformable attention.}
\section{Conclusions}
\label{sec:conclusions}

We introduced I-LW-DETR, a fully integer-only realization of LW-DETR for efficient object detection. By jointly redesigning the architecture and its numerical operators, we showed that competitive detection accuracy can be maintained under both PTQ and QAT despite eliminating all floating-point computations. The proposed SD-ShiftGELU and calibrated Shiftmax address the main numerical limitations of previous integer transformer approximations. The ablation studies confirm that these components effectively eliminate the principal quantization bottlenecks while preserving detection performance.

The experimental results demonstrate that I-LW-DETR achieves a favorable accuracy-efficiency trade-off among fully integer-only DETR models. 
After QAT, the proposed pipeline incurs only a 4.3--5.4 mAP degradation with respect to floating-point LW-DETR across all model scales, while reducing the model size by approximately $3.6\times$ and the computational cost by $12.9$--$14.4\times$. Experiments on VisDrone further demonstrate that the proposed methodology generalizes to the challenging task of small-object detection. Future work will investigate deployment on representative edge AI accelerators and evaluate runtime performance on real hardware.

\section{Acknowledgments}
This work was supported by the French National Research Agency (ANR) under the “Investissements d’avenir” program ANR 10-AIRT-0005 \& “France 2030” program ANR-26-AIRT-0004 (IRT NANOELEC).  % version finale

%\appendix
%\input{content/Pseudocode}

%\clearpage
\appendix

\section{\texorpdfstring{Pseudo-code of the proposed methods}
                         {Pseudo-code of the proposed methods}}

\label{app1}

\begin{algorithm}[H]
    \footnotesize
    \tcp{This algorithm implements integer-only SD-ShiftGELU. It first computes
    the integer exponential approximation using ShiftExp, then applies a
    sign-dependent stabilization before integer division to compute the GELU output.}
    \renewcommand{\baselinestretch}{1.1}\selectfont
    \caption{\textbf{SD-ShiftGELU}}
    \label{alg:sd_shiftGELU}
    \SetAlgoLined
    \vskip 0.05in

    \KwIn{
    \\
    $I_{x}$: Integer input \\
    $S_{x}$: Input scaling factor \\
    $k_{out}$: Output bit-precision \\
    $k_{inter}$: Intermediate bit-precision \\
    }

    \vskip 0.1in
    \SetKwFunction{FSHIFT}{ShiftExp}
    \SetKwProg{Fn}{Function}{:}{}
    \Fn{\FSHIFT{$I, S, k_{inter}$}}{
    $I_e \leftarrow I + (I \gg 1) - (I \gg 4)$
    \hfill {\color{gray}$\triangleright$ $I \cdot \log_2 e$} \\

    $I_0 \leftarrow \lfloor 1/S \rceil$ \\

    $I_e \leftarrow \texttt{clamp\_min}(I_e, k_{inter} \cdot (-I_0))$
    \hfill {\color{gray}$\triangleright$ Lower-bound clipping} \\

    $q \leftarrow \lfloor I_e / (-I_0) \rfloor$
    \hfill {\color{gray}$\triangleright$ Integer part} \\

    $r \leftarrow -(I_e - q \cdot (-I_0))$
    \hfill {\color{gray}$\triangleright$ Decimal part} \\

    $I_b \leftarrow ((-r) \gg 1) + I_0$ \\

    $I_{exp} \leftarrow I_b \ll (k_{inter} - q)$
    \hfill {\color{gray}$\triangleright$ Integer exponential approximation} \\

    $S_{exp} \leftarrow S / (2^{k_{inter}})$ \\

    \Return $(I_{exp}, S_{exp})$
    \hfill {\color{gray}$\triangleright$ $S_{exp}\cdot I_{exp}\approx e^{S \cdot I}$}
    }

    \vskip 0.1in
    \SetKwFunction{FGELU}{SD-ShiftGELU}
    \SetKwProg{Fn}{Function}{:}{}
    \Fn{\FGELU{$I_{x}, S_{x}, k_{out}, k_{inter}$}}{
    $I_p \leftarrow I_{x} + (I_{x} \gg 1) + (I_{x} \gg 3) + (I_{x} \gg 4)$
    \hfill {\color{gray}$\triangleright$ $1.702I_x$} \\

    $I_{\mathrm{shift}} \leftarrow \texttt{ReLU}(I_p)$
    \hfill {\color{gray}$\triangleright$ Sign-dependent shift} \\

    $I_{\Delta} \leftarrow I_p - I_{\mathrm{shift}}$
    \hfill {\color{gray}$\triangleright$ $I_{\Delta}\leq 0$} \\

    $(I_{exp}, S_{exp}) \leftarrow \texttt{ShiftExp}(I_{\Delta}, S_{x}, k_{inter})$ \\

    $(I_{exp}', S_{exp}') \leftarrow \texttt{ShiftExp}(-I_{\mathrm{shift}}, S_{x}, k_{inter})$ \\

    $(I_{div}, S_{div}) \leftarrow \texttt{IntDiv}(I_{exp}, I_{exp} + I_{exp}', k_{out})$
    \hfill {\color{gray}$\triangleright$ Sigmoid approximation} \\

    $(I_{out}, S_{out}) \leftarrow (I_{x} \cdot I_{div},\, S_{x} \cdot S_{div})$ \\

    \Return $(I_{out}, S_{out})$
    \hfill {\color{gray}$\triangleright$ $I_{out}\cdot S_{out} \approx \text{GELU}(I_{x}\cdot S_{x})$}
    }

    \vskip 0.05in
\end{algorithm}

\begin{algorithm}[H]
    \footnotesize
    \tcp{This algorithm implements constrained integer-only Shiftmax. It uses ShiftExp to approximate the exponential values, calibrates a denominator
    shift to keep the normalization within 32-bit integer precision, and uses
    quotient-remainder accumulation to preserve the exact shifted exponential sum.}
    \renewcommand{\baselinestretch}{1.1}\selectfont
    \caption{\textbf{%Hardware-Aware
    Constrained Shiftmax}}
    \label{alg:hardware_shiftmax}
    \SetAlgoLined
    \vskip 0.05in

    \KwIn{
    \\
    $I_x$: Integer attention-score vector \\
    $S_x$: Attention-score scaling factor \\
    $k_{out}$: Output bit-precision \\
    $k_{inter}$: Intermediate bit-precision \\
    $m_p$: Precision margin \\
    $M_{hw}$: Hardware-supported division precision, fixed to $31$ \\
    }

    \vskip 0.1in
    \SetKwFunction{FCALIB}{CalibrateShift}
    \SetKwProg{Fn}{Function}{:}{}
    \Fn{\FCALIB{$S_{sum}, m_p, M_{hw}$}}{
    $M_{req} \leftarrow \left\lceil \log_2(S_{sum}) \right\rceil$
    \hfill {\color{gray}$\triangleright$ Required denominator precision} \\

    $M_{target} \leftarrow M_{req} + m_p$
    \hfill {\color{gray}$\triangleright$ Add precision margin} \\

    $s_d \leftarrow \max(M_{target} - M_{hw}, 0)$ \\

    \Return $s_d$
    }

    \vskip 0.1in
    \SetKwFunction{FSOFT}{ConstrainedShiftmax}
    \SetKwProg{Fn}{Function}{:}{}
    \Fn{\FSOFT{$I_x, S_x, k_{out}, k_{inter}, m_p, M_{hw}$}}{
    $I_{max} \leftarrow \max(I_x)$ \\

    $I_{\Delta_i} \leftarrow I_{x_i} - I_{max}$
    \hfill {\color{gray}$\triangleright$ Ensure $I_{\Delta_i}\leq 0$} \\

    $(I_{exp_i}, S_{exp}) \leftarrow \texttt{ShiftExp}(I_{\Delta_i}, S_x, k_{inter})$ \\

    $S_{sum} \leftarrow \sum_i I_{exp_i}$
    \hfill {\color{gray}$\triangleright$ Accumulated exponential sum} \\

    $s_d \leftarrow \texttt{CalibrateShift}(S_{sum}, m_p, M_{hw})$ \\

    $Q_i \leftarrow I_{exp_i} \gg s_d$
    \hfill {\color{gray}$\triangleright$ Quotient of each exponential value} \\

    $r_i \leftarrow I_{exp_i} - (Q_i \ll s_d)$
    \hfill {\color{gray}$\triangleright$ Remainder of each exponential value} \\

    $S_R \leftarrow \sum_i Q_i + \left( \sum_i r_i \gg s_d \right)$
    \hfill {\color{gray}$\triangleright$ Exact shifted denominator} \\

    $S_R \leftarrow \texttt{clamp}(S_R, 1, 2^{31}-1)$ \\

    $F \leftarrow \left\lfloor \dfrac{2^{31}-1}{S_R} \right\rfloor$
    \hfill {\color{gray}$\triangleright$ 32-bit normalization factor} \\

    $I_{out_i} \leftarrow (I_{exp_i} \cdot F) \gg (31-k_{out}+1+s_d)$
    \hfill {\color{gray}$\triangleright$ Convert to $k_{out}$-bit output} \\
    
    $S_{out} \leftarrow 2^{-(k_{out}-1)}$
    \hfill {\color{gray}$\triangleright$ Fixed-point Softmax scale} \\
    
    \Return $(I_{out}, S_{out})$
    \hfill {\color{gray}$\triangleright$ $S_{out}I_{out_i}\approx \mathrm{Softmax}(S_xI_x)_i$}
    }

    \vskip 0.05in
\end{algorithm}

\clearpage

%\bibliographystyle{elsarticle-num} 
%\bibliography{main.bib}

%for ArXiv



\end{document}